\documentclass[10pt,journal,compsoc]{IEEEtran}
\usepackage[ruled,vlined,linesnumbered]{algorithm2e}
\usepackage{color,graphicx,epstopdf,changepage,amsmath,multirow,booktabs}
\usepackage[justification=centering]{caption}
\usepackage{blindtext}
\usepackage[inline]{enumitem}
\usepackage{blindtext}
\usepackage{xcolor}
\usepackage{flushend}
\usepackage{amsthm}
\usepackage{subcaption}
\usepackage{afterpage}
\usepackage{lscape}
\SetAlgoCaptionSeparator{.\space}

\theoremstyle{definition}
\newtheorem{exmp}{Example}[section]
\usepackage{tikz}

\makeatletter
\g@addto@macro\normalsize{%
  \setlength\abovedisplayskip{3pt}
  \setlength\belowdisplayskip{3pt}
  \setlength\abovedisplayshortskip{3pt}
  \setlength\belowdisplayshortskip{3pt}
  \setlength{\textfloatsep}{3pt plus 1.0pt minus 2.0pt}
  \setlength{\intextsep}{3pt}
}
\makeatother

\newcommand{\cse}{C}

\newcommand{\gra}{\mathcal{G}}

\renewcommand{\mathbf}[1]{{\color{green}{#1}}}

\newcommand{\pop}[1]{\mathcal P^{#1}}
\newcommand{\nhm}[1]{\mathcal {NHM}^{#1}}

\begin{document}
\title{Memetic EDA-Based Approaches to Comprehensive Quality-Aware Automated Semantic Web Service Composition}
\author{Chen~Wang, Hui~Ma, Gang~Chen, and~Sven~Hartmann
\IEEEcompsocitemizethanks{\IEEEcompsocthanksitem Chen~Wang, Hui~Ma, and~Gang~Chen are with the School of Engineering and Computer Science,  Victoria University of Wellington, Wellington,
New Zealand. \protect\\
E-mail: \{chen.wang, hui.ma, aaron.chen\}@ecs.vuw.ac.nz
\IEEEcompsocthanksitem Sven~Hartmann is with the Department of Informatics, Clausthal University of Technology, Germany.\protect\\
E-mail: sven.hartmann@tu-clausthal.de}
}
\markboth{}%
{Shell \MakeLowercase{\textit{et al.}}: Bare Demo of IEEEtran.cls for Computer Society Journals}
\IEEEtitleabstractindextext{%
\begin{abstract}
Comprehensive quality-aware automated semantic web service composition is an NP-hard problem, where service composition workflows are unknown, and comprehensive quality, i.e., Quality of services (QoS) and Quality of semantic matchmaking  (QoSM) are simultaneously optimized.  The objective of this problem is to find a solution with optimized or near-optimized overall QoS and QoSM within polynomial time over a service request.   In this paper, we proposed novel memetic EDA-based approaches to tackle this problem. The proposed method investigates the effectiveness of several neighborhood structures of composite services by proposing domain-dependent local search operators. Apart from that, a joint strategy of the local search procedure is proposed to integrate with a modified EDA to reduce the overall computation time of our memetic approach. To better demonstrate the effectiveness and scalability of our approach, we create a more challenging, augmented version of the service composition benchmark based on WSC-08 \cite{bansal2008wsc} and WSC-09 \cite{kona2009wsc}. Experimental results on this benchmark show that one of our proposed memetic EDA-based approach (i.e.,  MEEDA-LOP)  significantly outperforms existing state-of-the-art algorithms.
\end{abstract}

\begin{IEEEkeywords}
Web service composition, QoS optimization, Combinatorial optimization,
EDA.
\end{IEEEkeywords}}

\maketitle
\IEEEraisesectionheading{\section{Introduction}\label{sec:introduction}}
\IEEEPARstart{S}{ervice} Oriented Architecture (SOA) has been contributing to the reuse of software components \cite{booth2004web}. \emph{Web services} are one of the most successful implementations of SOA to provide services as ``modular, self-describing, self-contained applications that are available on the Internet" \cite{curbera2001web}. Often, users' requirements cannot be satisfied by one existing web service, \emph{Web service composition} aims to loosely couple a set of Web services to provide a value-added composite service (i.e., a solution of service composition) that accommodates users' complex requirements. These requirements are related to functional (i.e., quality of semantic matchmaking as QoSM) and non-functional (i.e., Quality of service as QoS) requirements, which give birth to \emph{semantic web service composition} and \emph{QoS-aware web service composition}, with the aim of optimizing QoSM and QoS of service composition solutions respectively. Many researchers have been working on solving these optimization problems in web service composition \cite{ma2015hybrid,peng2017estimation,rodriguez2010composition,da2016genetic,da2017evolutionary,wang2017comprehensive,wang2017gp,wang2018knowledge,yu2013adaptive}.

Existing works that study the above problems are classified as \emph{semi-automated} and \emph{fully-automated} web service composition \cite{rao2004survey} with two different assumptions. One assumes that users know an abstract service composition workflow, and all the composite services produced by the composition system must strictly obey the given workflow. However, this assumption is not always valid since the workflow may not be provided or not even known by users. The second group of research works does not rely on any existing workflows. Instead, a composite service will be constructed from scratch by selecting and connecting multiple atomic services obtained from the service repository  \cite{rao2004survey}. Therefore, this construction process can end up with different workflows. Apparently, compared to semi-automated web service composition, fully-automated web service composition opens new opportunities to improve QoS and QoSM further due to different workflows automatically constructed. Nevertheless, the difficulty of the composition task is also increased.

AI planning and Evolutionary Computation (EC) are two of the most widely used techniques for \emph{semi-automated} and \emph{fully-automated} web service composition \cite{ma2015hybrid,rodriguez2010composition,wang2017comprehensive,yu2013adaptive,peer2005web,qi2010combining,wang2018eda}. AI planning techniques focus on creating valid composite services, where the functional correctness is always ensured with gradually constructed workflows.  However, these approaches do not optimize the QoS or QoSM of the solutions produced \cite{tong2011distributed}. EC techniques have been widely used to solve service composite problems that aim to optimize either one or both of QoSM and QoS, and are potentially more useful in practice as they can efficiently find "good enough" composite solutions. Important approaches \cite{ma2015hybrid,peng2017estimation,rodriguez2010composition,da2016genetic,da2017evolutionary,wang2017comprehensive,wang2017gp,wang2018knowledge,yu2013adaptive} based on Genetic Algorithms (GA) \cite{whitley1994genetic}, Genetic Programming (GP) \cite{koza1992genetic}, Particle Swarm Optimization (PSO) \cite{kennedy1995particle} and Estimation of Distribution Algorithm (EDA) \cite{tsutsui2006comparative}, have been widely investigated in the literature. 

To effectively search for good solutions, EC techniques often employ useful information distilled from promising solutions to produce new offspring. The information can be used either implicitly or explicitly.  Conventional EC techniques, such as GA and GP,  fall in the implicit camp by producing new solutions through recombining solutions evolved previously \cite{ma2015hybrid,rodriguez2010composition,yu2013adaptive}. In contrast, one EC technique that has achieved prominent success through explicit use of information is Estimation of Distribution Algorithm (EDA) \cite{ceberio2012review}. In EDA, information about promising solutions evolved previously is captured compactly in the form of probability models. EDA has been successfully utilized for semi-automated service composition \cite{peng2017estimation,pichanaharee2008qos}, but they can not support fully automated service composition. We recently proposed a new EDA-based approach for fully automated web service composition through reliable and accurate learning of a probability model that encodes the distribution of promising solutions \cite{wang2018knowledge}, i.e., a distribution model.

EDA stresses more on global exploration, rather than local exploitation \cite{wang2016estimation}. It is due to that the distribution model has the objective of exploring more promising regions in the entire solution space, without attempting to improve the quality of any specific solutions evolved previously. However, the optimization performance can often be improved directly through local modifications to promising solutions. By restricting the target region for local search and avoiding most of the randomness involved in sampling directly from the distribution model, this can potentially expedite the search of optimal solutions. Therefore, to improve its competency in finding more effective solutions, an idea is to enhance EDA with local search, namely, memetic EDA. Memetic EDA has been successfully applied to many optimizations problems with local search operators \cite{wang2016estimationrouting,wang2016estimation}, such as arc routing and assembly flow-shop scheduling problems.

On the one hand, although memetic EDA has been successfully applied to many applications,  those memetic approaches work inappropriate for web service composition, as these local search operators are only applicable to domain-specific or problem-specific solution representations \cite{wang2016estimation,wang2016estimationstochastic}. On the other hand, despite the recent success in EDA-based service composition, the effectiveness of this approach can be enhanced by introducing memetic EDA. Several challenges remain to be addressed in developing a memetic EDA approach to service composition as follows:

First, a composite service is commonly represented as a DAG, exploring the neighborhood of a DAG, especially large DAGs, is computationally infeasible \cite{acid2003searching}. Note that the discussed neighborhood is structured by local search operators on the search space, where neighbor solutions can be generated  iteratively from a given candidate solution. Therefore, researchers \cite{da2017evolutionary,lacomme2004competitive} often indirectly defined the neighborhood of a composite service represented in the form of a permutation, which can be converted to a DAG through a separate decoding process. Often, so-called ``swap'' operators produce neighbors by swapping two random elements in a permutation. Consequently, a neighborhood is defined by the collection of permutations obtainable through a ``swap'' to any given permutation. However, such neighborhood often contains a large proportion of neighboring permutations with inferior quality. For effective local research, the neighborhood must be refined to exclude most of the clearly unwise swapping choices by exploiting domain-specific knowledge.

Second, as we know, it is very challenging to determine which candidate solutions are to be selected for local search in memetic algorithms, as the selection method has a significant impact on the effectiveness and efficiency of memetic EDA. Should an equal chance be given to all the candidate solutions or only elite solutions should be considered for local search? Moreover, what are elite solutions, and how many of them should be modified locally?   However, answers to these challenging questions often vary from many factors, such as EC algorithms,  domain problems, etc.  Therefore, it is challenging to determine one effective selection strategy for the memetic EDA-based approach to service composition. 

Third, a traditional strategy that exclusively explores the whole neighboring space of composite services can incur high computation cost without guarantee of improving solution quality. For example, for permutation-based representation, if a simple swap operator is utilized for exploring the neighborhood, then the dimension of the permutation determines the computational complexity. In the context of service composition, the dimension of such permutation is usually equivalent to the size of the service repository. As the neighborhood size is extremely huge when many services are to be considered during the service composition process, this strategy is infeasible for practical use.

Fourth, in EDA, although a probability distribution model is adjusted to trace promising searching areas throughout generations,  one proportion of promising solutions (i.e., permutations) are more likely to be repetitively sampled,  while the distribution model is getting converged along the generations. Furthermore, these repeatedly sampled solutions are often favorable to users, since they are candidate solutions with high quality. In the EDA-based approach to service composition, sampled permutation-based solutions are very costly as they require repetitive computation time for decoding and evaluations.

To address these challenges above, we propose a memetic EDA-based approach, achieving substantially high performances in effectiveness and efficiency. These outstanding performances are observed by comparing it with some recently proposed web service composition approaches, such as a EDA-based approach \cite{wang2018knowledge}, a PSO-based approach  \cite{wang2017comprehensive}, and GA- and Memetic GA-based approaches  \cite{da2017evolutionary}. In particular, an empirical, experimental study on the effectiveness of different neighborhoods structured by different local search operators is conducted. The contributions of this paper are listed below, and where the first contribution is to address the first challenge discussed previously, and the second contribution is proposed to address the remaining challenges.

\begin{enumerate}
\item To perform an effective local search in composite services, we first propose several neighborhood structures for candidate solutions. These neighborhoods are created by developing several novel domain-dependent local search operators, based on constructing and swapping effective building blocks of composite services for local improvements. Subsequently, we develop an effective memetic EDA-based approach based on our previous work \cite{wang2018knowledge}, with nature integration with those local search operators.

\item To significantly reduce the computation time of our proposed memetic EDA-based  approach, an integrated local search procedure is proposed with a modified EDA based on the standard EDA. To decrease computation losses in repetitive sampling and evaluations,  we utilize an archiving technique to avoid sampling solutions repetitively. This technique is prevalent and straightforward to use. Besides that, the local search procedure employs an effective joint strategy to efficiently finding better solutions. This strategy considers  fitness uniform distribution scheme and stochastic local search jointly with our proposed local search operators. 

\item To demonstrate the performance of our memetic EDA-based approach, we create a more challenging, augmented version of the service composition benchmark based on WSC-08 \cite{bansal2008wsc} and WSC-09 \cite{kona2009wsc}. In particular, the new benchmark inherits the functionalities provided by services in benchmark dataset WSC-08 and WSC-09 and the QoS attributes of web services in benchmark dataset QWS \cite{al2007qos}. Moreover, the number of web services in the service repository is doubled as a new benchmark (with much bigger searching space) to demonstrate that memetic EDA can maintain high performance on our problem with significantly larger sizes. This benchmark has been made freely available online as well as the codes of our memetic EDA-based approach \footnote{ Two augmented benchmarks for automated web service composition is available from https://github.com/chenwangnida/Dataset, and the codes of our memetic EDA-based approach is available from     https://github.com/chenwangnida/MENHBSA4SWSC. }. We experimentally compare our memetic EDA-based approach with some state-of-the-art methods that have been recently proposed to solve the same or a similar service composition problem using the new benchmark. Our experimental results illustrate that our method can achieve cutting-edge performance.
\end{enumerate}
\section{Related Work}\label{relatedwork}
In this section, we review some state-of-the-art EC-based service composition approaches for solving fully automated service composition. Afterwards,  we discuss some memetic EC-based approaches, where local search is introduced to enhance the performance. Lastly, we focus on discussing one promising EC-based algorithm, i.e.,  EDA, and some recent success that has been achieved by memetic EDA-based approaches for other problems. 
\subsection{Literature on EC-Based fully automated web service composition}
Automated web service composition aims to loosely couple web services to fulfill a  service request, without strictly obeying a pre-given abstract workflow. Instead, composition workflows are gradually built up while its component services are selected. Existing works in fully automated web service composition can be categorized into two approaches --- direct approaches and indirect approaches \cite{da2018evolutionary}. The direct approaches represent composition solutions explicitly in the representation that displays actual execution flows of composite services, while the indirect approaches often represent composite services implicitly as permutations, which require a decoding process to build up actual execution workflows.

In the first category, tree- and graph-based representations are widely used to represent service composition solutions directly. A graph-based evolutionary process is introduced in \cite{da2015graphevol} to directly evolve DAG-based service composition solutions, applying domain-dependent crossover and mutation operators with repairing methods. GP is utilized for searching optimal solutions represented as trees. \cite{rodriguez2010composition} proposes a context-free grammar for randomly initializing tree-based service composition solutions with correct structures of composite services. In contrast,  \cite{yu2013adaptive} randomly initializes tree-based service composition solutions completely, but they develop adaptive crossover and mutation rates according to the diversity of the population for accelerating the speed of convergence. Both approaches \cite{rodriguez2010composition,yu2013adaptive} utilize a penalization method for filtering incorrect solutions while evaluating the QoS of candidate solutions. To achieve higher performance, \cite{ma2015hybrid,da2016genetic} utilize a greedy search algorithm for creating correct DAG-based composition workflows, which are mapped to tree-based ones with different methods. During the evolutionary process, the correctness of the solutions is ensured by domain-dependent crossover and mutation. However, the mapped tree-based representations suffer a scalability issue, since many replicas of subtrees are produced from the mapping methods. To overcome this issue, \cite{wang2017gp} proposes a tree-like representation, on which the replicas of subtrees are handled by removing them, and inserting edges from the root of the replicas to the roots of the copies.

In the second category, service composition solutions are represented as permutations, which are then decoded into solutions represented as DAGs \cite{wang2017comprehensive,da2018evolutionary,da2016particle}. PSO is utilized to find an optimized queue of services (i.e., a permutation), which can be decoded into a corresponding DAG-based composite service  \cite{da2016particle}.  \cite{wang2017comprehensive} extends \cite{da2016particle} to jointly optimize QoSM and QoS, where a weighted DAG is decoded, where edge weights correspond to matchmaking quality between services. These two PSO-based approaches rely on PSO to determine the weights of particle's position (that corresponding with a service) to form an ordered service queue. Optimizing QoSM and QoS simultaneously is more challenging than optimizing QoS only because the searching space has significantly increased, and it demands more effective and efficient searching techniques. Apart from that, it has been suggested that utilizing the indirect representation often contributes to a higher performance, compared to direct representation \cite{da2018evolutionary}. It is due to that the search space is not unwittingly restricted by unconstrained random initialization of solutions and operators. 

In summary, EC techniques have been showing their promises in fully automated web service composition. Moreover, the indirect approaches have been indicated to be more effective. Therefore, EC techniques with indirect representations are exciting techniques to be focused on for solving service composition problem in this paper. 
\subsection{Literature on memetic EC-based approaches and EDA}
Memetic algorithms have drawn growing attention from researchers in recent years and achieved significant successes in many applications \cite{chen2011multi}. By introducing local search, the performance of EC techniques can be improved. In the domain of service composition,  to overcome the prematurity and proneness of GP, Tabu search is combined with GP to solve QoS-aware data-intensive web service composition \cite{yu2014hybrid}. \cite{da2017evolutionary} proposed an indirect memetic approach for QoS-aware web service composition, where a domain-dependent crossover operator is proposed to produce candidate solutions. Besides that, an exhaustive local search is applied to composite solutions represented as permutations. However,  the produced neighbors are likely to be  decoded into the same composite solution. Therefore, the effectiveness of this local search operator demands further improvement.  

Recently, EDA has been used as a technique to tackle permutation-based optimization problems \cite{ceberio2012review}. In particular,  a distribution model is learned iteratively for each population. Subsequently, new offsprings are generated based on the learned model. Moreover, domain-dependent local search operators are often introduced to enhance the performances of EDA. For example, a probability matrix that is related to the job priority permutation of a solution is learned in EDA-based flow-shop scheduling problem, and different job-based local search operators were proposed to enhance the exploitation ability of EDA  \cite{wang2016estimation}. An Edge Histogram Matrix is applied to uncertain capacitated arc routing problems and is leaned from solutions represented by a set of routes \cite{wang2016estimationstochastic}. To make local improvements,  different move operators, such as single insertion and swap, are also proposed. 

The use of EDA has only been investigated for semi-automated web service composition \cite{peng2017estimation,pichanaharee2008qos,mao2012empirical}. However, we recently proposed an EDA-based approach for fully automated web service composition, where candidate solutions are represented as permutations over a given service repository. The success of the proposed method strongly depends on the distribution model and the way of learning the distribution model. We employ Node Histogram Matrix (NHM) to learn the distribution of promising solutions in one population, Node Histogram-Based Sampling Algorithm (NHBSA) \cite{tsutsui2006comparative} is empoloyed to produce candidate solutions. Although we started an initial study for fully automated service composition, it remains an opportunity to improve its performance further. EDA is good at global exploration, and local search operators are motivated to be introduced in EDA to enhance its capability in exploitation. 

In summary, on the one hand, memetic EDA-based approaches have been investigated in many problems, other than fully automated service composition,  achieving promising results. On the other hand, notwithstanding success achieved in our initial investigation in EDA-based fully automated service composition, the performance of this EDA-based approach can be further improved by combining it with local search.
\section{Semantic Web Service Composition Problem}\label{Problem Description}
\begin{table*}
\footnotesize
\centering
\caption{QoS calculation for a composite service expression $\cse$}
\begin{tabular}{l|l|l|l|l}
\hline
 $\cse=$       &$r_\cse=$                              &$a_\cse=$                              &$ct_\cse=$                            &$t_\cse=$ \\ \hline
 $\bullet(\cse_1,\ldots,\cse_d)$      &$\prod\limits^d_{k=1}r_{\cse_k}$    &$\prod\limits^d_{k=1}a_{\cse_k}$    &$\sum\limits^d_{k=1}ct_{\cse_k}$   &$\sum\limits^d_{k=1}t_{\cse_k}$  \\ \hline
 $\parallel(\cse_1,\ldots,\cse_d)$  &$\prod\limits^d_{k=1}r_{\cse_k}$    &$\prod\limits^d_{k=1}a_{\cse_k}$    &$\sum\limits^d_{k=1}ct_{\cse_k}$   &$MAX \{ t_{\cse_k} | k \in \{ 1,...,d \} \}$\\ \hline
 $+(\cse_1,\ldots,\cse_d)$     &$\prod\limits^d_{k=1}p_k\cdot r_{\cse_k}$    &$\prod\limits^d_{k=1}p_k\cdot a_{\cse_k}$    &$\sum\limits^d_{k=1}p_k\cdot ct_{\cse_k}$   &$\sum\limits^d_{k=1}p_k\cdot t_{\cse_k}$  \\ \hline
 $\ast \cse_0$         &${r_{\cse_0}}^\ell$  &${a_{\cse_0}}^\ell$  &$\ell\cdot ct_{\cse_0}$ &$\ell\cdot t_{\cse_0}$ \\ \hline
\end{tabular}
\label{tbl:QoS_Aggre}
\end{table*}

A \emph{semantic web service} (\emph{service}, for short) is considered as a tuple $S =(I_{S}, O_{S}, $ $QoS_S)$ where $I_{S}$ is a set of service inputs that are consumed by $S$, $O_{S}$ is a set of service outputs that are produced by $S$, and $QoS_{S}=\{t_S, c_S, r_S, a_S\}$ is a set of non-functional attributes of $S$. The inputs in $I_{S}$ and outputs in $O_{S}$ are parameters modeled through concepts in a domain-specific ontology $\mathcal{O}$. The attributes $t_S, c_S, r_S, a_S$ refer to the response time, cost, reliability, and availability of service $S$, respectively, which are four commonly used QoS attributes \cite{zeng2003quality}.

A \emph{service repository} $\mathcal{SR}$ is a finite collection of services supported by a common ontology $\mathcal{O}$. A \emph{composition task} (also called \emph{service request}) over a given $\mathcal{SR}$ is a tuple $T=(I_{T}, O_{T})$ where $I_{T}$ is a set of task inputs, and $O_{T}$ is a set of task outputs. The inputs in $I_{T}$ and outputs in $O_{T}$ are parameters that are semantically described by concepts in the ontology $\mathcal{O}$.

Two special atomic services $Start = (\emptyset, I_T, \emptyset )$ and $End  = (O_T, \emptyset, \emptyset)$ are always included in $\mathcal{SR}$ to account  for the input and output of a given composition task $T$.

We use \emph{matchmaking types} to describe the level of a match between outputs and inputs \cite{paolucci2002semantic}. For concepts $a, b$ in $\mathcal{O}$ the \emph{matchmaking} returns $exact$ if $a$ and $b$ are equivalent ($a \equiv b$), $plugin$ if $a$ is a sub-concept of $b$ ($a \sqsubseteq b$), $subsume$ if $a$ is a super-concept of $b$ ($a \sqsupseteq b$), and $fail$ if none of previous matchmaking types is returned. In this paper we are only interested in $exact$ and $plugin$ matches for robust compositions, see \cite{lecue2009optimizing}. As argued in \cite{lecue2009optimizing} $plugin$ matches are less preferable than $exact$ matches due to the overheads associated with data processing. For $plugin$ matches, the semantic similarity of concepts is suggested to be considered when comparing different $plugin$ matches.

A \emph{robust causal link} \cite{lecue2008optimizing} is a link between two matched services $S$ and $S'$, denoted as $S \rightarrow S'$, if an output $a$ ($a \in {O_S}$) of $S$ serves as the input $b$ ($b \in {O_{S'}}$) of $S'$ satisfying either $a \equiv b$ or $a \sqsubseteq b$.  For concepts $a, b$ in $\mathcal{O}$, the \emph{semantic similarity} $sim(a, b)$ is calculated based on the edge counting method in a taxonomy like WorldNet \cite{shet2012new}. Advantages of this method are simple calculation and good semantic measurement \cite{shet2012new}. Therefore, the \emph{matchmaking type} and \emph{semantic similarity} of a robust causal link is defined as follows:

\begin{equation}
\footnotesize
\label{eq_type}
type_{link} = 
\begin{cases}
	1 & \text{ if $a\equiv b$ ($exact$ match)}\\
	p & \text{ if $a \sqsubseteq b$ ($plugin$ match)}
\end{cases}
\end{equation}

\begin{equation}
\footnotesize
\label{eq_sim}
sim_{link} = sim(a, b) = \frac{2N_c}{N_{a}+N_{b}}
\end{equation}

\noindent with a suitable parameter $p$, $0< p < 1$, and with $N_a$, $N_b$ and $N_c$, which measure the distances from concept $a$, concept $b$, and the closest common ancestor $c$ of $a$ and $b$ to the top concept of the ontology $\mathcal{O}$, respectively. However, if more than one pair of matched output and input exist from service $S$ to service $S'$, $type_e$ and $sim_e$ will take on their average values.

The \emph{QoSM} of a composite service is obtained by aggregating over all robust causal links as follows:

\begin{equation}
\footnotesize
\label{eq_TYPE}
MT {=} \prod_{j=1}^{m} type_ {link_{j}}
\end{equation}

\begin{equation}
\footnotesize
\label{eq_SIM}
SIM {=} \frac{1}{m}\sum_{j=1}^m sim_ {link_{j}}
\end{equation}

Formal expressions as in \cite{ma2012formal} are used  to represent service compositions. The constructors $\bullet$, $\parallel$, $+$ and $\ast$ are used to denote sequential composition, parallel composition, choice, and iteration, respectively. The set of \emph{composite service expressions} is the smallest collection $\mathcal{SC}$ that contains all atomic services and that is closed under sequential composition, parallel composition, choice, and iteration. That is, whenever $\cse_0,\cse_1,\ldots,\cse_d$ are in $\mathcal{SC}$ then $\bullet(\cse_1,\ldots,\cse_d)$, $\parallel(\cse_1,\ldots,\cse_d)$, $+(\cse_1,\ldots,\cse_d)$, and $\ast \cse_0$ are in $\mathcal{SC}$, too. Let $\cse$ be a composite service expression. If $\cse$ denotes an atomic service $S$ then its QoS is given by $QoS_S$.  Otherwise the QoS of $\cse$ can be obtained inductively as summarized in Table~\ref{tbl:QoS_Aggre}. Herein, $p_1,\ldots,p_d$ with $\sum\limits^d_{k=1}p_k=1$ denote the probabilities of the different options of the choice $+$, while $\ell$ denotes the average number of iterations. Therefore, QoS of a service composition solution, i.e., availability ($A$), reliability ($R$),  execution time ($T$), and cost ($CT$)  can be obtained by aggregating $a_\cse$, $r_\cse$, $t_\cse$ and $ct_\cse$ as in Table~\ref{tbl:QoS_Aggre}.

In the presentation of this paper, we mainly focus on two constructors, sequence $\bullet$ and parallel $\parallel$, similar as in most automated service composition works \cite{ma2015hybrid,da2016genetic,wang2017comprehensive,wang2017gp,da2015graphevol,da2016particle}, where service composition solutions are represented as a Directed Acyclic Graph (DAG). We can easily calculate  QoS of a composite service that is represented as a DAG \cite{wang2017comprehensive} according to Table~\ref{tbl:QoS_Aggre}.
 
When multiple quality criteria are involved in decision making, the fitness of a solution is defined as a weighted sum of all individual criteria in Eq.~(\ref{eq:fitness}), assuming the preference of each quality criterion based on its relative importance is provided by the user \cite{hwang1981lecture}:

\begin{equation}
\footnotesize
\label{eq:fitness}
Fitness (\cse) = w_1 \hat{MT} + w_2 \hat{SIM} + w_3 \hat{A} + w_4 \hat{R} + w_5(1 - \hat{T}) + w_6(1 - \hat{CT})
\end{equation}

\noindent with $\sum_{k=1}^{6} w_k= 1$. This objective function is defined as a  \emph{comprehensive quality model} for service composition. We can adjust the weights according to the user's preferences. $\hat{MT}$, $\hat{SIM}$, $\hat{A}$, $\hat{R}$, $\hat{T}$, and $\hat{CT}$ are normalized values calculated within the range from 0 to 1 using Eq.~(\ref{eq_normal}). To simplify the presentation we also use the notation $(Q_1,Q_2,Q_3,Q_4,Q_5,Q_6) $ $= (MT,SIM,A,R,T,CT)$. $Q_1$ and $Q_2$ have minimum value 0 and maximum value 1. The minimum and maximum value of $Q_3$, $Q_4$, $Q_5$, and $Q_6$ are calculated across all the relevant services (that are determined in  Sect.~\ref{subsection:initilization}) in the service repository $\mathcal{SR}$ using greedy search in \cite{ma2015hybrid,da2016genetic}.

\begin{equation}
\footnotesize
\label{eq_normal}
\hat{Q_k} = 
\begin{cases}
	\frac{Q_k - Q_{k, min}}{Q_{k, max} - Q_{k, min}} & \text{ if $k=1,\ldots,4$ and }Q_{k, max} - Q_{k, min} \neq 0,\\
	\frac{Q_{k,max} - Q_k}{Q_{k, max} - Q_{k, min}} & \text{ if $k=5,6$ and }Q_{k, max} - Q_{k, min} \neq 0,\\
	1 & \text{ otherwise}.
\end{cases}
\end{equation}

\noindent The goal of comprehensive quality-aware service composition is to find a composite service expression $\cse^{\star}$ that maximizes the objective function in Eq.~(\ref{eq:fitness}). $\cse^{\star}$ is hence considered as the best possible solution for a given composition task $T$. 

\begin{figure*}[h!tb]
\footnotesize
\centering
\includegraphics[width=0.85\textwidth]{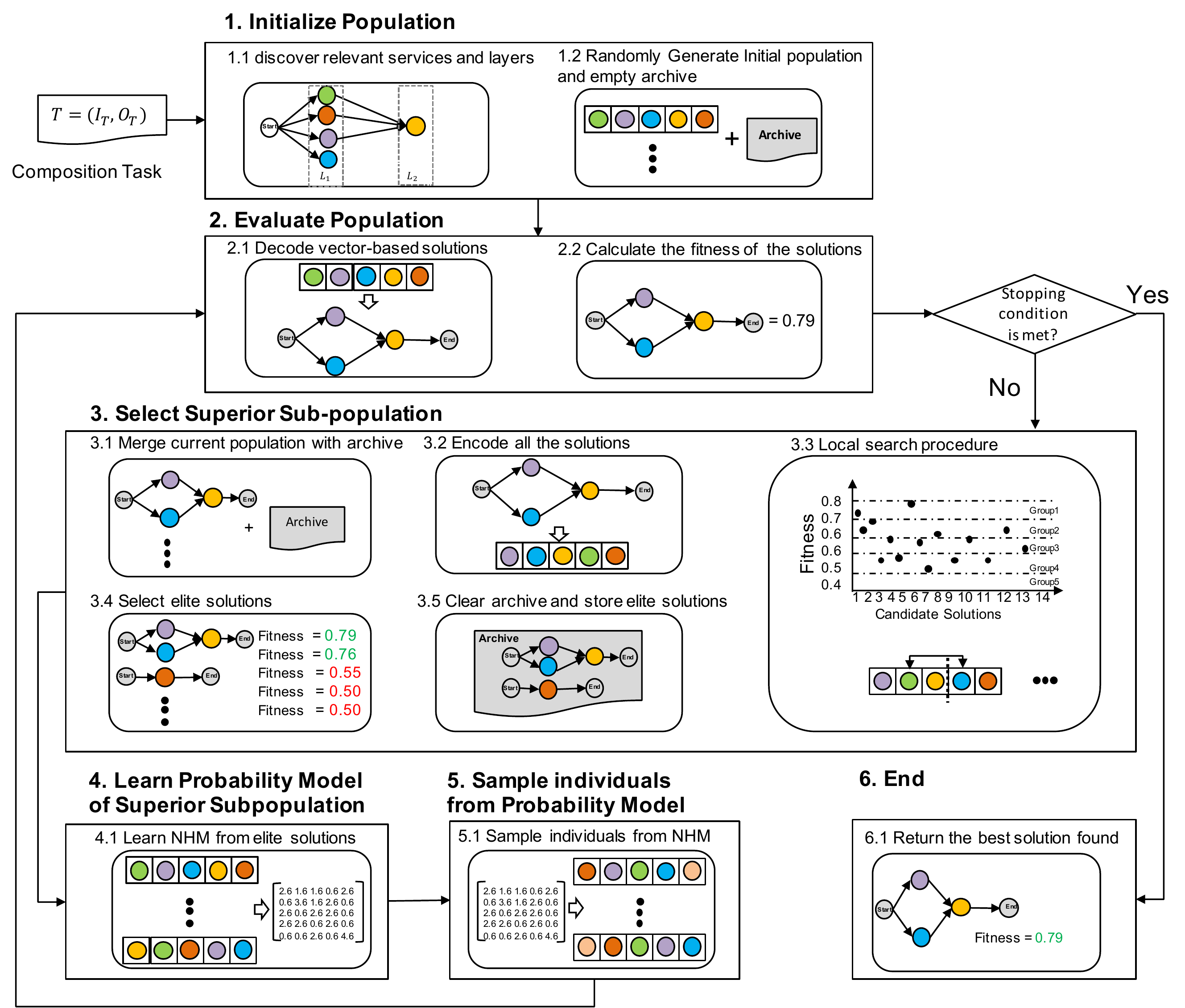}
 \caption{An overview of memetic EDA-based approach for \\automated web service composition }
 \label{fig:overview}
\end{figure*}
\section{Memetic EDA-Based Approach for Semantic Web Service Composition}\label{eda_appraoch}
In this section, we present our memetic EDA-based approach to fully automated semantic web service composition. We start by giving an overview of our memetic EDA-based approach. Subsequently, we discuss some essential steps in the approach: the first one is to discover relevant services and service layers, see details in Sect.\ref{subsection:initilization}. The second one is to introduce a permutation-based representation proposed in our previous work, see details in Sect. \ref{subsection:representation} and \ref{subsection:NHM}. The third one is to introduce an effective joint strategy for a local search procedure, see details in Sect. \ref{subsection:local search}.

We propose several key ideas that are jointly employed to build our memetic EDA-based approach:
\begin{enumerate}
\item A composite service is commonly represented as a DAG, since a DAG can intuitively represent an execution flow of web services, and allows efficient computation of QoS. The success of the EDA strategy strongly relies on the proper distribution model for learning the knowledge of promising solutions. Our initial study \cite{wang2018knowledge} represents a composite service as a unique queue of services, i.e., a permutation of atomic services, which is mapped from a DAG-based solution. Composite services in this permutation form contributes to a distribution model to be learned and new permutation-based promising solutions to be sampled.  Therefore, a bi-directional map is ensured between permutations and DAGs for learning and evaluation purposes.
\item To significantly decrease the computation time of the local search procedure, it is crucial to select a restricted number of suitable candidate solutions for local searches. We assume that candidate solutions with close fitness values are similar in their corresponding DAG forms, so neighbors produced from these candidate solutions can be the same.  Therefore, we group candidate solutions based on their fitness values according to a uniform distribution scheme, which allows candidate solutions with the most considerable differences measured by single-objective fitness values can be effectively chosen for applying local search.  
\item It is not efficient to exhaustively explore the whole neighbors in the conventional local search \cite{da2017evolutionary}. Instead, stochastically searching the neighboring solutions can significantly reduce computation cost  \cite{wang2016estimationrouting}. Therefore, we introduce a stochastic local search with EDA to efficiently exploit the neighborhood of the selected candidate composite services. 
\item Exploring neighborhood of a large DAG-based composite service is unusually computationally infeasible \cite{acid2003searching}.  However, it is straightforward to define the neighborhood on a permutation-based representation by so-called swap operators. To develop effective swap operators,  we utilize domain knowledge of service composition to create effective building blocks for these swap operators on permutation-based candidate solutions. These swap operators aim to exploit fitter neighbors effectively. That is they are likely to make local improvements in the produced neighbors. 
\end{enumerate}
\subsection{An overview of memetic EDA-based algorithm for automatic service composition}
An overview of the memetic EDA-based approach is represented in Figure~\ref{fig:overview}, consisting of the following steps: initialize population, evaluate population, select superior sub-population, learn probability model, sample individuals and return optimal solutions. We start with discovering all the relevant services that are related to a given composition request $T$ in Step 1. Meanwhile, several service layers are identified (see details in Subsection~\ref{subsection:initilization}). These relevant services are used to randomly generate $m$ composite services represented as permutations, ${\Pi^{\prime}}^{g}_k$, where $g= 0$ and $k=1,\dots,m$. In Step 2, these permutation-based individuals are decoded into DAG-based solutions using a forward graph building technique \cite{wang2017comprehensive}, based on which, the fitness in Eq.~\ref{eq:fitness} of each individual can be calculated. In Step 3, we merge the current population $\pop{g}$ with an archive. The archive is an empty individual set initially and will saved with elite composite services in the future. By adopting Breath-First Search (BFS) on each corresponding DAG-based solutions in the merged population, we produce another encoded permutation-based solutions $\Pi^{g}_k$. Then, the local search procedure is applied to a very small set of these permutations. This small permutation set is selected based on a fitness uniform selection scheme over the current population (see details in \ref{scheme}). For each permutation in the small set, a stochastic local search is  employed to create new permutations as its neighbors, where the best neighbor is identified based on the fitness value. This permutation in the small set is replaced with its best neighbor (see details in Subsection \ref{subsection:local search}). The top half of the best-performing solutions are reserved in $\pop{g}$ according to their fitness values and put them into the archive as elite solutions. In Step 4,  we use these elite solutions in the archive to learn a $NHM^g$ of generation $g$, which produces offsprings for generation $g+1$ using NHBSA, see details in Subsection~\ref{subsection:NHM}. Consequently,  we go back to Step 2 to evaluate the fitness of new offsprings. The steps 2 to 4 will be repeated until the maximum number of generations is reached. Eventually, the best solutions found throughout the evolutionary process is returned. 

In a nutshell, we introduce a permutation-based representation derived from the common DAG-based one. In our proposed algorithm, we always switch between these two representations back and forth for better searching or evaluation purposes. Furthermore, an effective and efficient local search procedure is developed through the use of the selection scheme and the  stochastic local search.

\subsection{Relevant Services and Service Layers}\label{subsection:initilization}
Discovering relevant services and service layers is an initial, but crucial step for our memetic EDA-based approach. We achieve two goals at this initial stage: the first goal is to reduce the size of the service repository $\mathcal{SR}$ to keep only those that are relevant to the service composition task $T$; the second goal is to identify service layers of these relevant services.  In particular, a group of layers is identified, and each layer contains a set of services that have the same longest distance to $Start $. We adopt a layer discovering method in \cite{da2017fragment} to find relevant services and service layers as illustrated in the following example. 


\begin{exmp}
\label{eg:relevant service}
We consider a \emph{composition task}  $T=(\{a, b\},\{ i\})$ and a $\mathcal{SR}$ consisting of seven atomic services. $S_0=(\{b \}, \{i\}, QoS_{S_0})$, $S_1=(\{a \}, \{f, g \}, QoS_{S_1})$, $S_2=(\{a, b\}, \{h \}, QoS_{S_2})$, $S_3=(\{f, h\}, \{ i\}, QoS_{S_3})$,  $S_4=(\{a\}, \{f, g, h\}, QoS_{S_4})$, $S_5=(\{a, c\}, \{f, g, h\}, QoS_{S_4})$ and $S_6=(\{c, d, e\}, \{f, g, h\}, QoS_{S_4})$. The two special services $Start=(\emptyset,\{a,b,e\},\emptyset)$ and $End=(\{i\},\emptyset,\emptyset)$ are defined by the given composition task $T$. Fig.~\ref{fig:decoding} shows an example of discovering relevant services and service layers given a service request $T$, where five related services (i.e., $S_0$, $S_1$, $S_2$, $S_3$, and $S_4$) and two layers (i.e., $\mathcal{L}_{1}$ and $\mathcal{L}_{2}$) are found. In $\mathcal{L}_{1}$, $S_0$, $S_1$, $S_2$, and $S_4$ can be satisfied by $\{a, b\}$ of $T$, and they have the same distance to $Start$ (Note that the distance is measured by the number of predecessors). While $S_3$ in $\mathcal{L}_{2}$ requires additional inputs from other services and it is associated with a longer distance to $Start$.
\end{exmp}

\begin{figure}[h!tb]
\footnotesize
\centering
\includegraphics[width=0.35\textwidth]{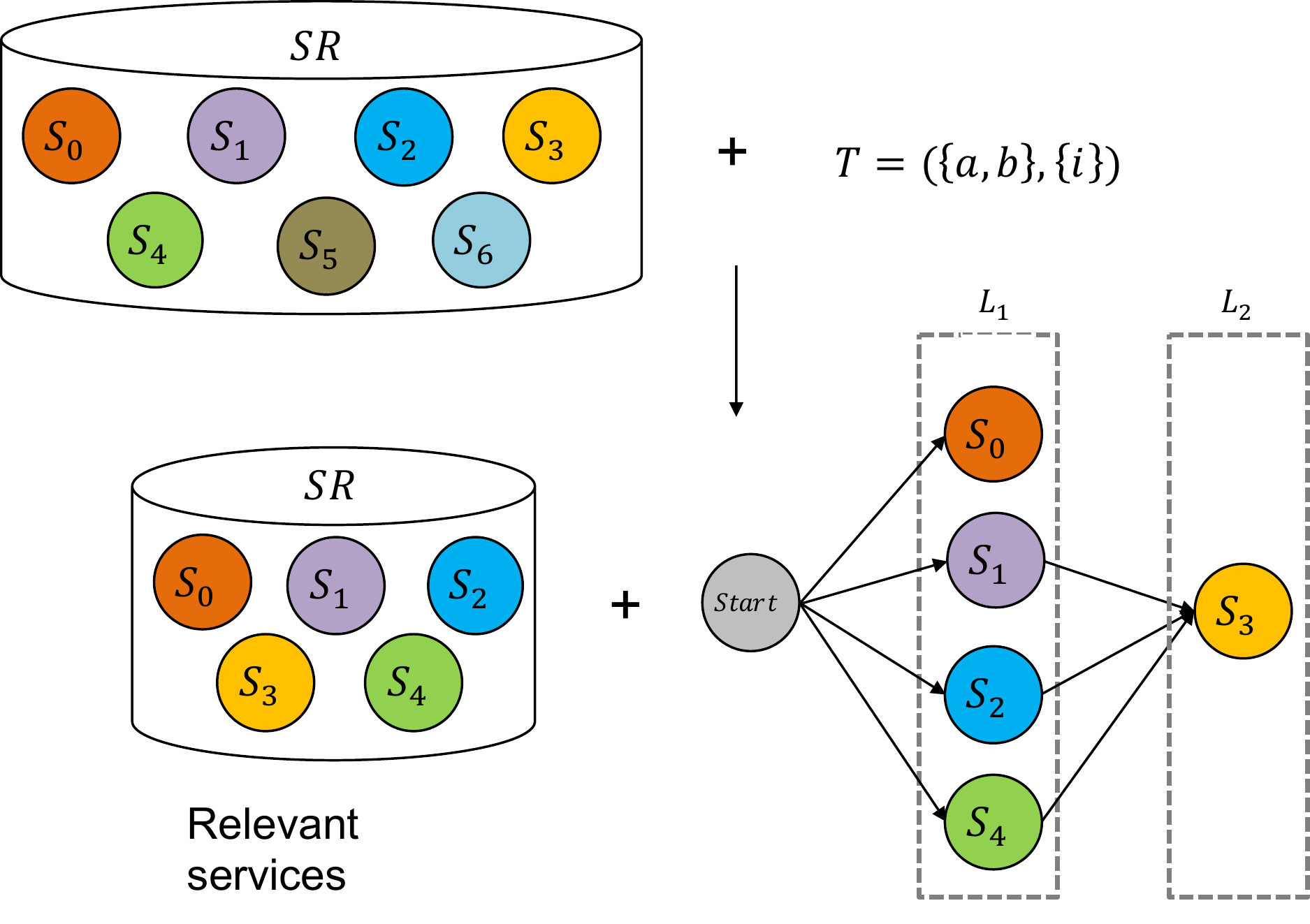}
 \caption{An example of discovering relevant services and \\ service layers for a service request $T$}
 \label{fig:two-point swap}
\end{figure}

 
\subsection{A Novel Permutation-Based Representation}\label{subsection:representation}
Service composition solutions are commonly represented as Directed Acyclic Graphs (DAGs) \cite{ma2015hybrid,da2016genetic,wang2017comprehensive,wang2017gp,da2015graphevol,da2016particle}. Let $\gra=(V, E)$ be a DAG-based composite solution from $Start$ to $End$, where nodes correspond to the services and edges correspond to the robust causal links. Often, $V$ does not contain all services in $\mathcal{SR}$.

Many combinatorial optimization problems naturally represent solutions as permutations, which can be different in different problems \cite{ceberio2012review}. Here we present composite services as permutations, and we ensure a bi-directional map between permutations and DAGs. The bi-directional map is crucial for learning the distribution of promising composite solutions. Because it is less reliable to learn a distribution based on permutations if different permutations are mapped to the same DAG-based composition service. Let $\Pi = (\Pi_0, \dots, \Pi_t,\Pi_{t+1}, \dots, \Pi_{n-1})$ be a permutation, elements of which are $\{ 0, \dots, t, t+1, \dots, n-1\}$ such that $\Pi_i \neq \Pi_j$ for all $i \neq j$. Particularly, $\{ 0, \dots, t\}$ are service indexes (i.e., id number) of the component services in the corresponding $\gra$ , and  is sorted based on the longest distance from $Start$ to every component services of  $\gra$. While $\{t+1, \dots, n-1\}$  be indexes of  remaining services in $\mathcal{SR}$ not utilized by  $\gra$. We use $\Pi^{g}_k$ to present  the $k^{th}$ (out of $m$, $m$ is population size) service composition solution, and $\pop{g} = [ \Pi_{0}^{g},  \ldots, \Pi_{k}^{g},\ldots, \Pi_{m-1}^{g} ]$ to represent a population of solutions of generation $g$. An example of producing a permutation-based composite solution is shown as follows.

\begin{exmp}
\label{example:1}
Let us consider a composition task $T=(\{a, b\},\{ i\})$ and a service repository $\mathcal{SR}$ consisting of five  services. $S_0=(\{b \}, \{i\}, QoS_{S_0})$, $S_1=(\{a \}, \{f, g \}, QoS_{S_1})$, $S_2=(\{a, b\}, \{h \}, QoS_{S_2})$, $S_3=(\{f, h\}, \{ i\}, QoS_{S_3})$ and $S_4=(\{a\}, \{f, g, h\}, QoS_{S_4})$. The two special services $Start=(\emptyset,\{a,b,e\},\emptyset)$ and $End=(\{i\},\emptyset,\emptyset)$ are defined by a given composition task $T$.  Fig.~\ref{fig:decoding} illustrates a process to produce an permutation-based solution. 

As an example, take an permutation as $[4, 1, 2, 3, 0]$. This service index queue is decoded into a DAG $\gra_0^0$ representing a service composition that satisfies the composition task $T$. Afterwards $\gra_0^0$ is mapped to a permutation $\Pi_{0}^{0} = [1, 2, 3$ $|$ $4, 0]$. Herein, each position on the left side of $|$ corresponds to a service discovered by a BFS on $\gra_0^0$ from $Start$. This BFS additionally takes ascending order of service indexes during the search. While the right side corresponds to the remaining atomic services in $\mathcal{SR}$, but not in $\gra_0^0$. Note, that $|$ is just displayed for the courtesy of the reader, rather than being part of the permutation-based representation. Furthermore, we also do not permit the encoding $[1, 2, 3$ $|$ $0, 4 ]$, as no information can be extracted from $\gra_0^0$ to determine the positions of $0$ and $4$ in the permutation. 

\begin{figure}[h!tb]
\footnotesize
\centering
\includegraphics[width=0.48\textwidth]{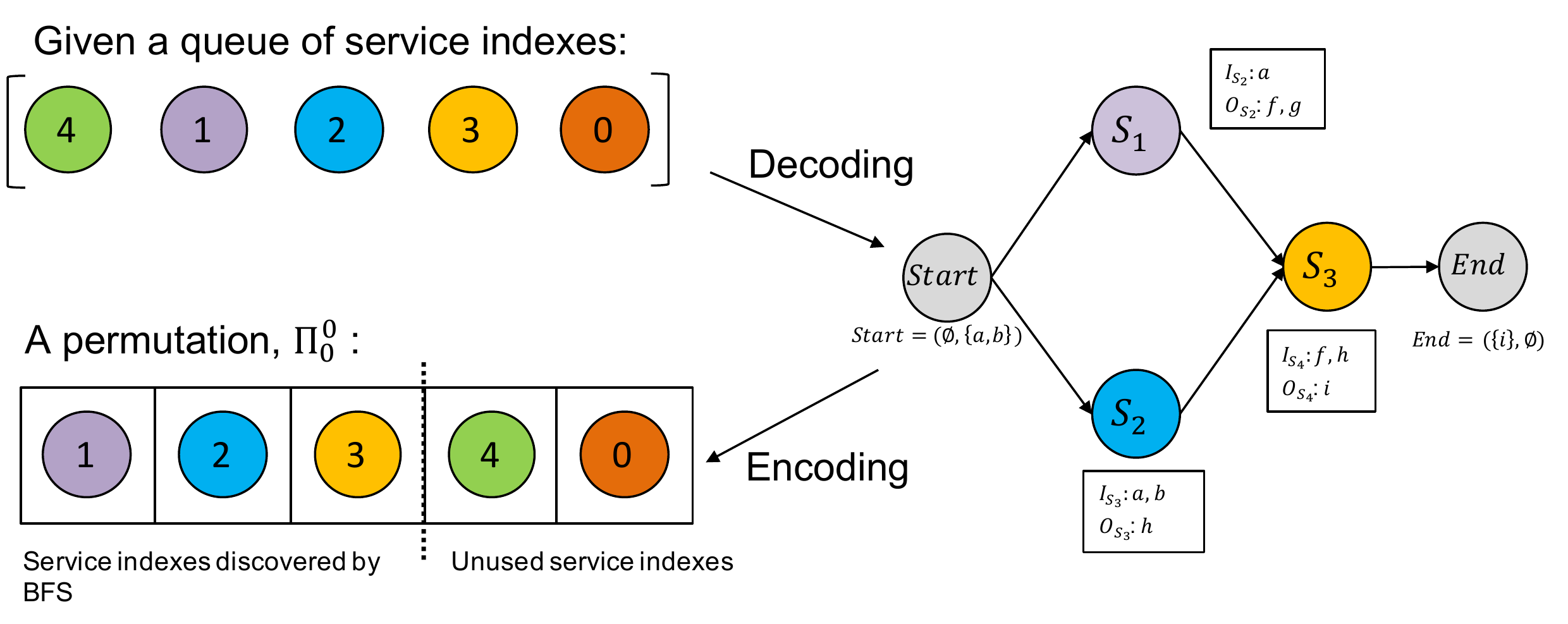}
 \caption{An example of $\gra$ over service request $T$}
 \label{fig:decoding}
\end{figure}

A permutation-based population $\pop{g}$ can be created with $m$ permutation-based solutions. Consider $m = 6$, $\pop{g}$ could be represented as follows:

\begin{equation*}
\pop{g} = 
\begin{bmatrix}
sol_{0}^{g}\\
sol_{1}^{g}\\
sol_{2}^{g}\\
sol_{3}^{g}\\
sol_{4}^{g}\\
sol_{5}^{g}
\end{bmatrix}
=
\begin{bmatrix}
1 & 2 & 3 & | &0 & 4 \\
0 & | & 1 & 2 & 3 & 4 \\
0 & | & 1 & 2 & 3 & 4 \\
4 & 3 & | & 0 & 1 & 2 \\
4 & 3 & | & 0 & 1 & 2 \\
2 & 1 & 3 &| & 0 & 4
\end{bmatrix}
=
\begin{bmatrix}
1 & 2 & 3 &0 & 4 \\
0 & 1 & 2 & 3 & 4 \\
0 & 1 & 2 & 3 & 4 \\
4 & 3 & 0 & 1 & 2 \\
4 & 3 & 0 & 1 & 2 \\
2 & 1 & 3 & 0 & 4
\end{bmatrix}
\end{equation*}
\end{exmp}
\subsection{Application of node histogram-based sampling}\label{subsection:NHM}
\cite{tsutsui2006comparative} proposed Node histogram-based sampling (NHBSA) as a tool for sampling new candidate solutions, which is commonly represented in the form of permutations. By employing the discussed representation of composite services in Sect.~\ref{subsection:representation}, we are now capable of applying NHBSA to sample new permutations as candidate composite services.

The NHM at generation $g$, denoted by $\nhm{g}$, is an $n\times n$-matrix with entries $e_{i,j}$ as follows:

\begin{equation}
\label{eq:NHM}
e^{g}_{i,j} = \sum_{k=0}^{n-1} \delta_{i,j}(sol_k^{g}) +  {\varepsilon} 
\end{equation}

\begin{equation}
\label{eq:delta}
\delta_{i,j} (sol_k^{g}) = 
\begin{cases}
	1 & \text{ if $I^{g}_{k}(S_{i})$ = j }\\
	0 & \text{ otherwise }
\end{cases}
\end{equation}

\begin{equation}
\label{eq:bias}
\varepsilon = \frac{m}{n-1} b_{ratio}
\end{equation}

\noindent where $i,j = 0, 1, \ldots, n-1$, and $b_{ratio}$ is a predetermined bias. Roughly speaking, entry $e^{g}_{i,j}$ counts the number of times that service $S_i$ appears in position $j$ of the service queue over all solutions in population $\pop{g}$.

\begin{exmp}
Consider $\pop{g}$ in Example \ref{example:1}, the size of population $m$ equals 6 and the dimension size of each individual (i.e., permutation) $n$ equals  5, and $b_{ratio}$ = 0.2, we calculate $\nhm{g}$ as follows:
\begin{equation*}
\nhm{g} = 
\begin{bmatrix}
2.6 & 1.6 & 1.6 & 0.6 & 2.6 \\
0.6 & 3.6 & 1.6 & 2.6 & 0.6 \\
2.6 & 0.6 & 2.6 & 2.6 & 0.6 \\
2.6 & 2.6 & 0.6 & 2.6 & 0.6 \\
0.6 & 0.6 & 2.6 & 0.6 & 4.6
\end{bmatrix}
\end{equation*}
\end{exmp}

We pick up an element in the $\nhm{g}$ as an example to demonstrate the meaning of each element in the NHM. For example,  $e^{g}_{0,0} $( that equals $2.6$) is made of integer and decimal parts: $2$ and $0.6$. The integer number $2$ means that service $S_0$ appears at the first position 2 times, while the decimal number 0.6 is a $\varepsilon$ bias. 

Once we have computed $\nhm{g}$, we use node histogram-based sampling \cite{tsutsui2006comparative} to sample new permutations for the next generation.

\subsection{Effective Local Search Procedure Through a Joint  Strategy}\label{subsection:local search}
In this section, we introduce a joint strategy of our local search procedure: we begin with an introduction of a selection of suitable individuals for employing local search. This selection aims to choose the individuals based on global and local population information using a fitness uniform selection scheme in Algorithim~\ref{alg:scheme}. Subsequently, we present several local search operators with the representation discussed in \ref{subsection:representation}. These operators are specially designed to work seamlessly with different neighborhoods that are investigated in this paper. The joint strategy for local search is summarized in ALGORITHM~\ref{alg:procedure}.

\begin{algorithm}
 \SetKwInOut{Input}{Input}\SetKwInOut{Output}{Output}
 \SetKwFunction{}{}
 \SetKwProg{Procedure}{Procedure}{}{}
 \Input{$\pop{g}$, $n_{nb}$ and $n_{set}$}
 \Output{updated $\pop{g}$}
 	Select a small number $n_{set}$ of individulals to form a subset $SelectedIndiSet$ of $\pop{g}$ using ALGORITHM~\ref{alg:scheme}\;
    \ForEach{$\Pi$ in $SelectedIndiSet$}{
    	Generate a size $n_{nb}$ of neighbors from $\Pi$ by local search \;
		Identify the best neighbor $\Pi_{best}$ with the highest fitness \;
        replace $\Pi$ with $\Pi_{best}$\;
    }
 \KwRet $\pop{g}$\;
 \caption{Joint  strategy for local search  (Step 3.3 in Fig.~\ref{fig:overview})}
\label{alg:procedure}
\end{algorithm} 

ALGORITHM~\ref{alg:procedure} takes three inputs: $\pop{g}$ the $g$th population, $n_{set}$ the number of seleted individuals for local search and $n_{nb}$ the number of neighbors. In this algorithm, we start by selecting a fixed and small number $n_{set}$ of candidate solutions  to form a subset $SelectedIndiSet$ of the current population $\pop{g}$ using ALGORITHM~\ref{alg:scheme}, see details in Section~\ref{scheme}. These selected solutions are used for local search. For each solution $\Pi$ in $SelectedIndiSet$, we produce a number $n_{nb}$ of neighbors from $\Pi$ by local search, see details in Section~\ref{operators}, and then we identify the best neighbor $\Pi_{best}$ from the produced neighbors. We replace the best neighbor $\Pi_{best}$ with the selected $\Pi$ in the small solutions set $SelectedIndiSet$. Eventually, we return a updated $\pop{g}$.

\subsubsection{Application of uniform distribution schema}
\label{scheme}
Two types of selection schemes for selecting suitable individuals for local search have been studied \cite{chen2011multi}: random selection scheme, and statistics scheme. The random selection scheme is a primary selection method, where a local search is potentially applied to all individuals with a pre-defined rate. However, it can be less effective as it does not assign local search to the most suitable candidate solutions, and it is more time-consuming when the population size is huge. This statistics scheme often chooses more suitable individuals based on the statistics information of the current population. For example, this method can assign local search on a set of candidate solutions with the highest differences measured by their fitness values.

\begin{algorithm}
 \SetKwInOut{Input}{Input}\SetKwInOut{Output}{Output}
 \SetKwFunction{createWeightedDAG}{createWeightedDAG}
 \SetKwProg{Procedure}{Procedure}{}{}
 \LinesNumbered
 \Input{$\pop{g}$ and $n_{set}$}
 \Output{ selected solutions $SelectedIndiSet$}
 $SelectedIndiSet \leftarrow \{  \}$ \;
 Sort $\pop{g}$ in descending order based on the fitness \;
 Put the first individual in $\pop{g}$ into $SelectedIndiSet$ \;
 Calculate fitness range for $n_{set}-1$ groups based on an uniform $interval$ between $maxfitness$ and $minfitness$ \;
 Assign each permutation in $\pop{g}$ to $n_{set}-1$ groups based on the fitness value \;
 Random select one permutation from each group and put it in $SelectedIndiSet$\;
 \KwRet $SelectedIndiSet$\;
 \caption{Fitness uniform selection scheme}
\label{alg:scheme}
\end{algorithm} 

Our selection scheme, inspired by \cite{huy2009adaptive}, is proposed based on the statistics information that aims to select a small number of suitable individuals for local search, making a good balance of local improvement and execution time.  This selection scheme is presented in ALGORITHM \ref{alg:scheme}. This algorithm applied  a local search on a set of selected individuals $SelectedIndiSet$. The size of $SelectedIndiSet$, $n_{set}$, is a pre-defined parameter. $SelectedIndiSet$ consists of one elite individual and $n_{set}-1$ individuals from $n_{set}-1$ groups of individuals in each generation. Particularly, we calculate a uniform fitness interval based on the maximal fitness value, $maxfitness$ and minimal fitness value,  $minfitness$ of the current population $\pop{g}$. Therefore, the population is divided into $n_{set}-1$ groups based on the calculated fitness interval. Consequently, these groups represent different groups of individuals, and each group represents close similarities based on their fitness. Note that, for every generation, the actual number of selected individuals for local search could be less than $n_{set}$, because there could be no individuals fall into one group based on its fitness value.

\subsubsection{Stochastic Local Search Operators}\label{operators}
To investigate an appropriate structure of neighborhood for composite services, suitable local search operators must be proposed to effectively utilize domain knowledge. Then we repeatedly assign these local search operators to $SelectedIndiSet$ for exploring their neighboring solutions. Apart from that, to balance the quality of local improvement and computation time, only a random subset of the entire large neighborhood is explored by a stochastic strategy. Based on the discussed permutation-based representation in Sect.~\ref{subsection:representation}, local search operators are proposed in a straightforward way as ``swap''. In this paper, we investigate four different swap operators:
\begin{enumerate}
\item \textbf{Constrained One-Point Swap}: For a permutation $\Pi = (\Pi_0, \dots, \Pi_t,\Pi_{t+1}, \dots, \Pi_{n-1})$, two service indexes $\Pi_a$, where $0 \leq a \leq t$, and $\Pi_b$, where $t+1 \leq b \leq n-1$, are selected and exchanged. 

The one-point swap local search operator is inspired by \cite{da2017evolutionary}, which swaps a pair of service indexes in a permutation. In \cite{da2017evolutionary},  local search exclusively explores the neighborhood based on one selected index of the permutation, so the size of the neighborhood associated with the index is $n-1$. However, it can be very computational expensive because the number of swaps becomes significant for large $n$. Besides that, it can be less flexible as the neighborhoods are just focusing on those neighborhoods in relation to one selected index. 

Herein we propose a more efficient and flexible local search with one-point swap:  first, we pre-determine a fixed, relatively small number of neighbors $n_{nb}$ to be produced for a considerable computational time assigned for local search; second, we randomly produce  $n_{nb}$ neighbors by swapping two randomly selected indexes, rather than by swapping $n-1$  indexes with one fixed index. We expect that swapping two randomly  selected indexes is more effective within a budget computation time for making local improvements. Meanwhile, we constrain the two randomly selected  indexes that they must be before $|$ and after $|$ respectively in every swap because these swaps exclude those have lower opportunities for local improvements. For example, one neighbor is created by swapping one pair of used service indexes. This swap operation has a higher chance to produce the same DAG-based solution. Figure \ref{fig:one_point_swap} shows an example of one-point swap for a selected individual. 

\begin{figure}
\footnotesize
\centering
\includegraphics[width=0.25\textwidth]{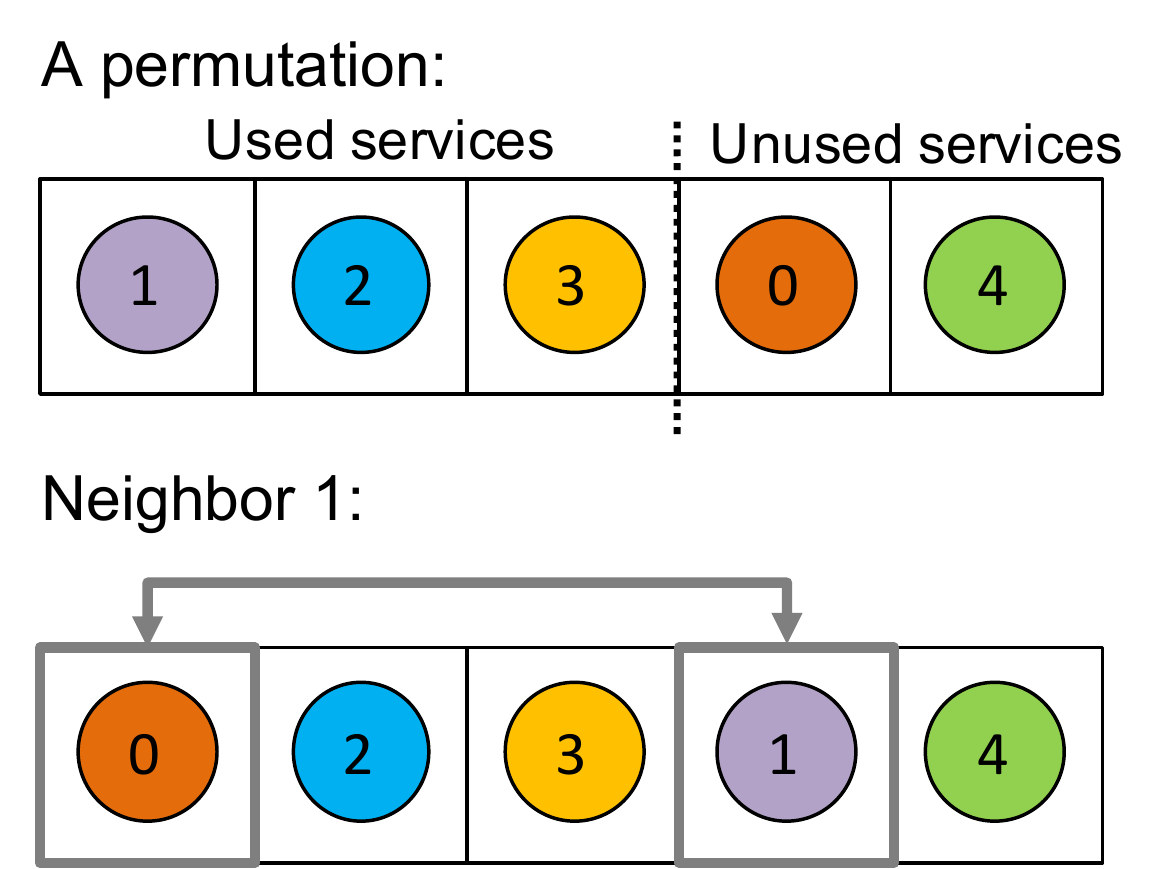}
 \caption{Examples of tone-point swap}
 \label{fig:one_point_swap}
\end{figure}

\item \textbf{Constrained Two-Point Swap}: For a permutation $\Pi = (\Pi_0, \dots, \Pi_t,\Pi_{t+1}, \dots, \Pi_{n-1})$, four service indexes $\Pi_{a_1}$, $\Pi_{a_2}$, $\Pi_{b_1}$, and $\Pi_{b_2}$ are selected, where $0 \leq a_1 \leq t$, $0 \leq a_2 \leq t$, $t+1 \leq b_1 \leq n-1$, $t+1 \leq b_2 \leq n-1$, $a_1 \neq a_2$, and $b_1 \neq b_2$. $\Pi_{a_1}$ and $\Pi_{b_1}$ are exchanged. Likewise, $\Pi_{a_2}$ and $\Pi_{b_2}$ are exchanged. 

Motivated by the one-point swap proposed above, we created two-point swap operator by combing two constrained one-point swap into a single operator. We make a hypothesis that the two-point swap could efficiently produce a higher quality neighbor by one local change, rather than producing two neighbors by a sequence of two constrained one-point local changes. Primarily, given a budgeted number of candidate solutions for local search, a two-point swap operator can perform a more efficient local search for finding high-quality solutions. Figure \ref{fig:two-point swap} shows an example of a two-point swap for a selected individual and a produced neighbors.

\begin{figure}[h!tb]
\footnotesize
\centering
\includegraphics[width=0.25\textwidth]{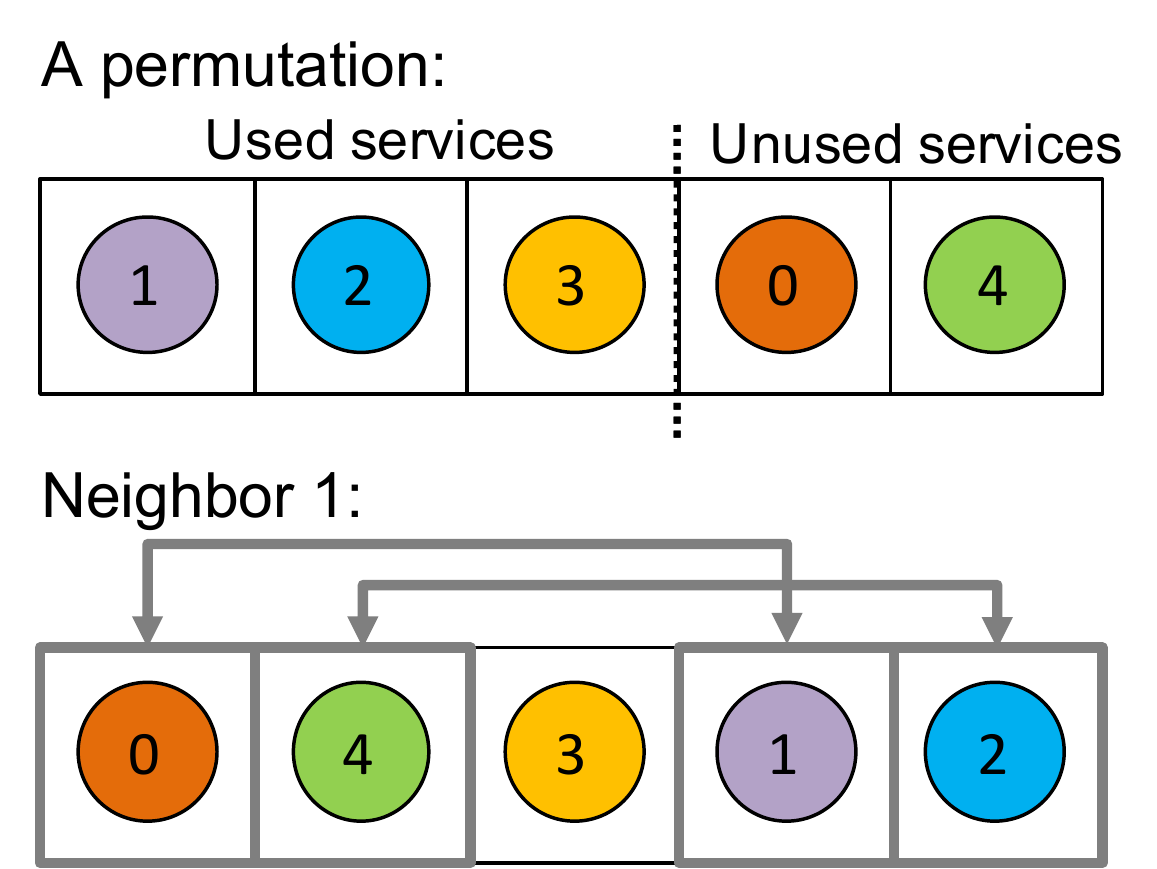}
 \caption{Examples of two-point swap}
 \label{fig:two-point swap}
\end{figure}

\item \textbf{Constrained One-Block Swap}: For a permutation $\Pi = (\Pi_0, \dots, \Pi_t,\Pi_{t+1}, \dots, \Pi_{n-1})$, two sub-blocks $\{ \Pi_a, \ldots, \Pi_t \}$, where where $0 \leq a < t $ and  $\{ \Pi_b, \ldots, \Pi_{n-1} \}$, where where $t+1 \leq b < n-1 $, are selected and exchanged.

Constrained One-Block Swap is proposed based on the concept of a block, i.e., consecutive points (service indexes) in a permutation. In this swap, two blocks are built up based on two randomly generated starting point $\Pi_a$ and $\Pi_b$ before $|$ and after $I$  of a permutation respectively.  After swaps, produced neighbors inherit two parts of the original permutation. Figure \ref{fig:two-point swap} shows an example of a constrained one-block swap for a permutation, where one block is built up from the start position $StartPos1$ to the last positions of used services, and another block is built up from the start position $StartPos2$ to the last index.

\begin{figure}[h!tb]
\footnotesize
\centering
\includegraphics[width=0.25\textwidth]{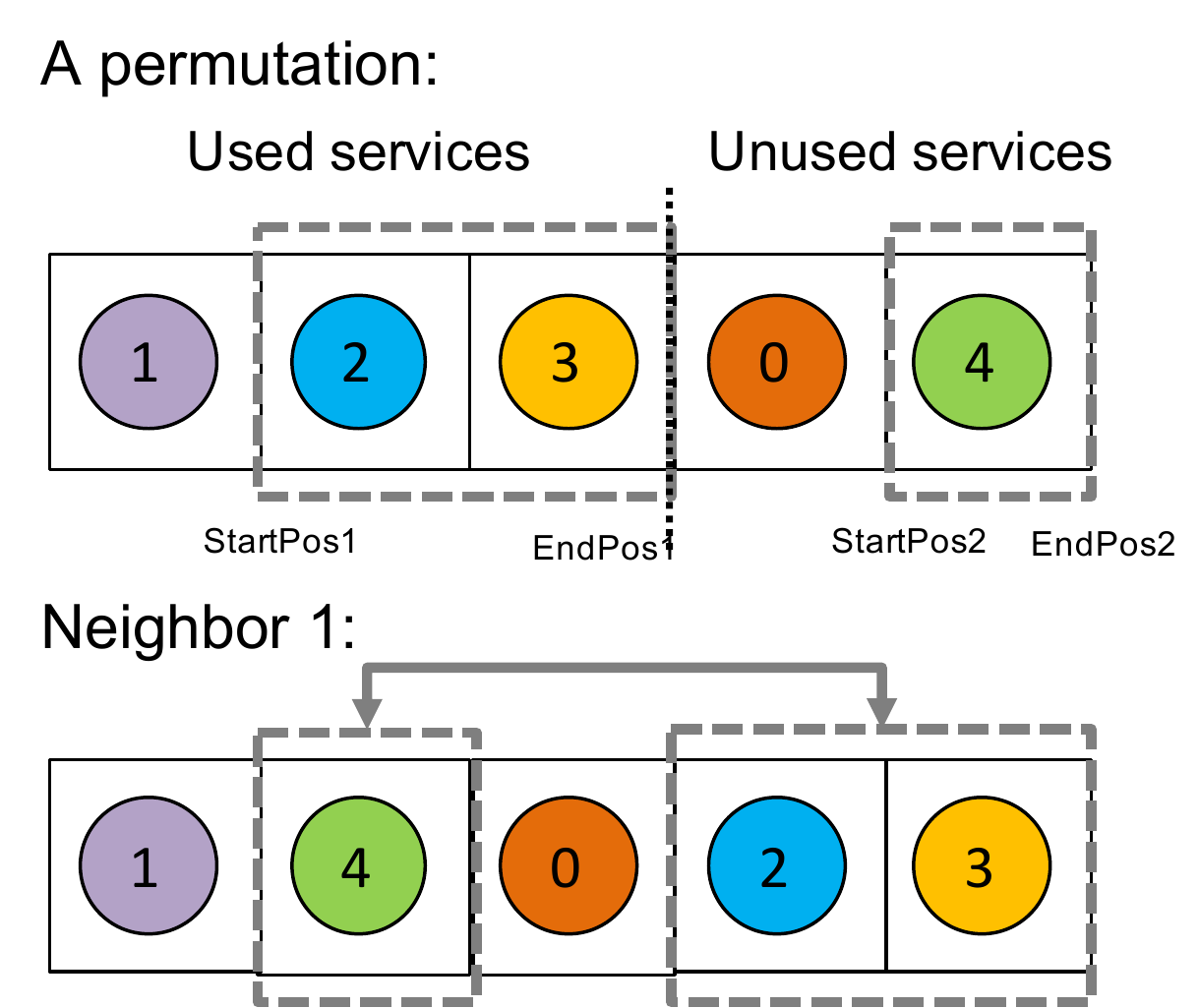}
 \caption{Example of Block swap operation}
 \label{fig:Block swap}
\end{figure}

\item \textbf{Layer-Based Constrained One-Point Swap}: For a permutation $\Pi = (\Pi_0, \dots, \Pi_t,\Pi_{t+1}, \dots, \Pi_{n-1})$, one service index  $\Pi_a$, where $0 \leq a \leq t$,  are selected, and one layer  $\mathcal{L}^{\prime}$, where $\mathcal{L}^{\prime}$ $s.t.$ $\Pi_a \in \mathcal{L}^{\prime}$, is identified. Afterwards, another service index $\Pi_b$ is randomly selected from the index set  $\mathcal{L}^{\prime} \cap \{\Pi_{t+1}, \dots, \Pi_{n-1}\}$. Consequently,  $\Pi_a$ and $\Pi_b$ are exchanged. 

Layer-based one-point swap operator is proposed by extending our one-point swap in additionally considering the layer information of relevant services. The layer information includes a set of layers, each of which estimates a set of services that could be located at positions of themselves in permutations. Herein we propose a layer-based one-point swap operator: we first select one service index and identify its associated layer, and then select another service index randomly from a set of indexes, i.e., an intersection of the indexes of the identified layer and the indexes of unused services. Consequently,  two service indexes are exchanged. Figure \ref{fig:block swap} shows an example of layer-based one-point swap for creating one neighbor from a selected individual.

\begin{figure}[h!tb]
\footnotesize
\centering
\includegraphics[width=0.45\textwidth]{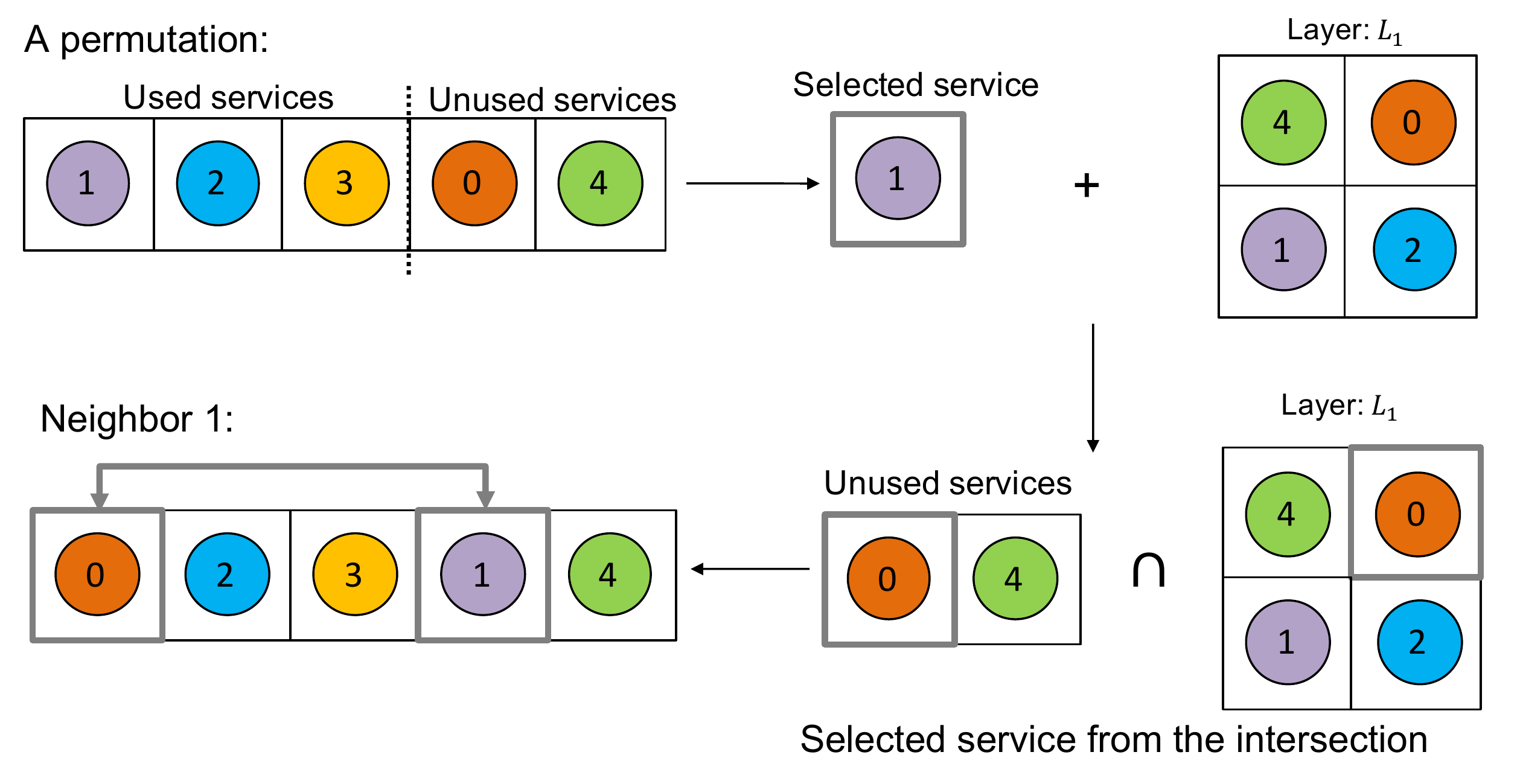}
 \caption{Example of layer-based one-point swap operation}
 \label{fig:block swap}
\end{figure}

\end{enumerate}

\section{Experiments}\label{section:experiments}
We conduct experiments to evaluate the performances of our memetic EDA-based approaches, i.e., memetic EDA with constrained one-point swap (henceforth referred to as MEEDA-OP),  memetic EDA with constrained two-point swap (henceforth referred to as MEEDA-TP), memetic EDA with constrained layer-based one-point swap (henceforth referred to as MEEDA-LOP) and memetic EDA with constrained one-block swap (henceforth referred to as MEEDA-OB).  These memetic EDA-based approaches are compared to some state-of-the-art EC-based methods that were recently proposed to solve the same or similar problems: a PSO-based approach \cite{wang2017comprehensive} (henceforth referred to as PSO), a GA-based approach (henceforth referred to as GA), a memetic GA-based approach \cite{da2017evolutionary} (henceforth referred to as MEGA) and an EDA-based approach \cite{wang2018knowledge} (henceforth referred to as  NHM-EDA). Two benchmarks, WSC-08 \cite{bansal2008wsc} and WSC-09 \cite{kona2009wsc} extended with QoS attributes , which generated from the QoS distribution from QWS \cite{al2007qos} are created. These two benchmarks have already been broadly employed in service composition  \cite{ma2015hybrid,wang2017comprehensive,yu2013adaptive} for experimental evaluations. Moreover, the number of web services in the service repository is doubled as a new benchmark (with much bigger searching space) to demonstrate that memetic EDA can maintain high performance on our problem with significantly larger sizes. We also make this benchmark available to the public. Particuarly, WSC08 contains 8 composition tasks with increasing size of service repository, i.e., 316, 1116, 1216, 2082, 2180, 4396, 8226, and 16238, and WSC09 contains 5 composition tasks with increasing size of service repository, i.e., 1144, 8258, 16276, 16602, and 30422 $\mathcal{SR}$s respectively.

The population size is set to 200, the number of generations equals to 100, and $b_{ratio}$ is 0.0002. The size of $SelectedIndiSet$ is 6, and the number of neighbors of each individual in $SelectedIndiSet$ explored by local search operators $n_{nb}$ is 20. For all the competing methods, we follow strictly their settings in their papers. In GA, the crossover rate is set to 0.95, and the mutation rate is set to 0.05. In MEGA, the crossover rate is set to 0.95, and local search rate is 0.05. We run the experiment with 30 independent repetitions. Following existing works \cite{wang2017comprehensive,wang2017gp, wang2018knowledge}, the weights of the fitness function Eq.~(\ref{eq:fitness}) are simply configured to balance the QoSM and QoS. In particular, we set both $w_1$ and $w_2$ to 0.25, and $w_3$, $w_4$, $w_5$ and $w_6$ all to 0.125. More experiments have been conducted and show that all our methods work consistently well under different weight settings. The $p$ of $type_{link}$ is determined by the preference of users, and is recommended as 0.75 for the plugin match according to \cite{lecue2009optimizing}. 


\subsection{Comparison of the Fitness}
We employ the independent-sample T-test with a significance level of 5\% to verify the observed differences in performance concerning fitness value and execution time. In particular, we use a pairwise comparison to compare all competing approaches, and then the top performances are identified, and its related value is highlighted in green color in Table~\ref{tbl:meanFitness}. Note that those methods that consistently find the best-known solutions over 30 runs with 0 standard deviations are also marked as top performances. The pairwise comparison results for fitness are summarized in Table~\ref{tbl:tableTallyFitness}, where textit{win}/\textit{draw}/\textit{loss} shows the scores of one method compared to all the others, and displays the frequency that this method outperforms, equals or is outperformed by the competing method. This testing and comparison methods are also used in Sect~\ref{section:time}.

One of the objectives of the experiments is to evaluate the effectiveness of the proposed memetic EDA-based approaches comparing to NHM-EDA \cite{wang2018knowledge}, PSO \cite{wang2017comprehensive}, GA and MEGA \cite{da2017evolutionary}. Table~\ref{tbl:meanFitness} shows the mean value of the fitness value and the standard deviation over 30 repetitions. The pairwise comparison results of the fitness value are summarized in Table~\ref{tbl:tableTallyFitness}. From Table~\ref{tbl:meanFitness} and Table~\ref{tbl:tableTallyFitness}, we observe some interesting behaviors of these approaches in finding high-quality solutions. Based on these observations, we also make some analysis and possible conclusions below:

Firstly, for the two baseline methods --- PSO and GA, all EDA-based approaches (with and without local search) consistently outperform PSO. However, only memetic EDA-based approaches outperform GA. 

Then, MEGA \cite{da2017evolutionary} achieved very comparable results to all our memetic EDA-based methods. However, MEEDA-LOP achieves the best performance. As shown in Table~\ref{tbl:tableTallyFitness},  MEEDA-LOP only loss 1 out of 13 composition tasks over WSC-08 land WSC-09. Furthermore,  MEEDA-LOP has achieved extremely stable performance in the most runs with 0 standard deviation. 

In addition, MEEDA-OP, MEEDA-TP, MEEDA-OB, and MEEDA-LOP significantly outperforms NHM-EDA \cite{wang2018knowledge}. This observation corresponds well with our expectation that the exploitation ability of EDA can be enhanced by hybridizing it with local search. We can see that all memetic EDA-based  approaches reach a better balance of exploration and exploitation. 

Furthermore, for the four memetic EDA-based approaches, MEEDA-OB is the worst while MEEDA-OP and MEEDA-TP are very comparable to each other. This observation demonstrates that the neighborhood based on blocks is considered to be less suitable for service composition problems,  it is due to that swapping building blocks can potentially ruin the learned distribution of promising solutions. 

Lastly, MEEDA-LOP is the best performer. This observation corresponds well with our assumption that using the layer-based information can further improve the effectiveness of one-point swap.  MEEDA-LOP applies the local search operator to a much smaller, but useful set of services considered in MEEDA-OP. 

In summary, we sort all the competing approaches based on the effectiveness in a descending order: MEEDA-LOP $>$ MEGA $>$  MEEDA-TP $=$ MEEDA-OP $>$ MEEDA-OB $>$ GA $>$ EDA $>$ PSO. 
\subsection{Comparison of the Execution Time}\label{section:time}
The second objective of our experiment is to study the efficiency of all the proposed EDA-based approaches comparing to EDA\cite{wang2018knowledge}, PSO \cite{wang2017comprehensive}, GA and MEGA \cite{da2017evolutionary}. Table~\ref{tbl:meanTime} shows the mean value of the execution time and the standard deviation over 30 repetitions. The pairwise comparison results for the execution time are summarized Table~\ref{tbl:tableTallyTime}. From the two tables above, we make some analysis and  possible conclusions about the execution time of these approaches as below:

First, MEEDA-LOP requires consistently less execution time compared to other approaches, which can be observed from the highlighted execution time in Table~\ref{tbl:meanTime}. It is a remarkable observation that the local search in MEEDA-LOP based on layers and constrained one-point swap requires less computation time compared to MEEDA-OP. However, this significant improvement is mainly due to two techniques in MEEDA-LOP. The first one is the archive technique, which reserves half population-size elite individuals to the next generation, and significantly reduces the overall computation time for the decoding and evaluation of the reserved individuals in the future. The second one is the layer-based information, which improves the effectiveness of one-point swap, resulting in learning more accurate and reliable NHM. Therefore, useful services are more likely to be put in the front of the permutation, which accelerates the execution time in the decoding process. 



\begin{figure}[!b]
\caption{A comparison of the average convergence rate of MEEDA-OP, MEEDA-TP, MEEDA-BP, MEEDA-LOP, NHM-EDA, PSO, MEGA and GA  over execution time on WSC08-3 (the left) and WSC09-2 (the right)}
\footnotesize
\centering
\includegraphics[width=0.45\textwidth]{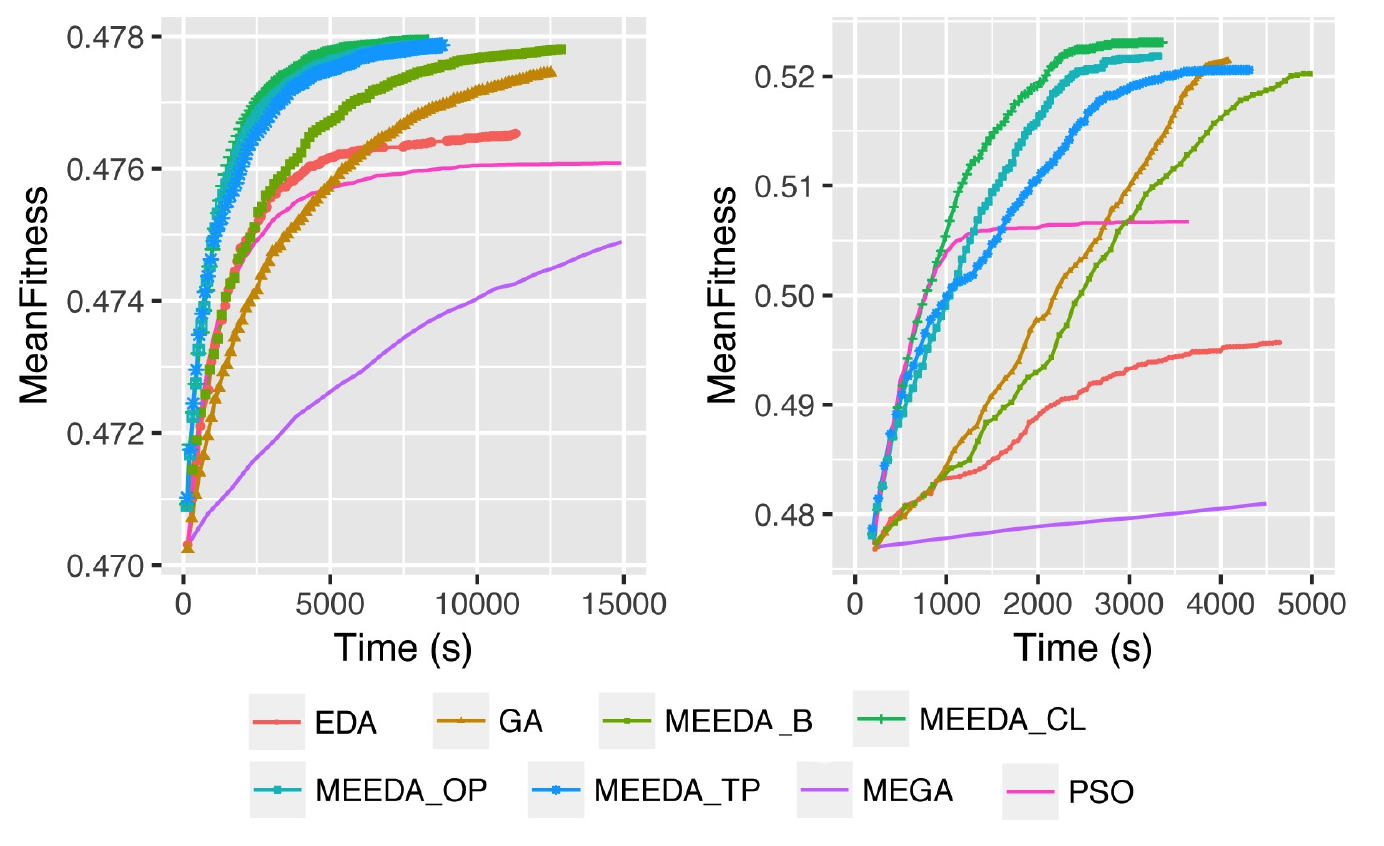}
\label{fig:convergency3}
\end{figure}

Second, in contrast,  MEGA requires the highest execution time,  because all the candidate solutions in MEGA have an opportunity for local search using random selection scheme, and MEGA also exclusively searches the whole neighborhood based on one position. These results confirm that the combination of the random selection scheme and the  exclusively local search strategy in MEGA is less effective and more time-consuming than our statistics scheme and stochastic local search operators. 

Lastly, MEEDA-OB is also very computation-intensive among all the memetic EDA-based approaches.  It is due to that one-block swap retards accurate distributions to be learned as local improvements of one-block swap is less effective, so required services for service composition are less likely to be put at the front of a service queue. Also, building blocks consume extra time in MEEDA-OB. 

In summary, we sort all the competing approaches based on the execution time in a ascending order:  MEEDA-LOP $>$ MEEDA-OP $>$ MEEDA-TP $>$ PSO $>$ GA $>$ MEEDA-OB $>$ MEGA.

\subsection{Comparison of the Convergence Rate }\label{subsection:convergence}
The third objective of our experiment is to study the convergence rate of all the approaches over 30 independent runs. We have used WSC08-3 and WSC09-2 as two examples to illustrate the performance of all the compared methods.

Fig.~\ref{fig:convergency3} exhibits the evolution of the mean fitness value of the best solution found along the execution time over 30 independent runs for MEEDA-OP, MEEDA-TP, MEEDA-OB, MEEDA-LOP,  NHM-EDA, PSO, GA, and MEGA. As MEGA requires much higher time for execution,  we set different execution time scales for two tasks of WSC08-3 and WSC09-2 to easily observe their differences.

First, we observe a significant increase in the fitness value towards the optimum over all the approaches excluding MEGA.  These approaches eventually reach different levels of plateaus. Given the same budget of execution time, all memetic EDA-based methods happen to converge significantly faster and require much less time than the baseline PSO over all the composition tasks. 

Second, MEGA suffers from the the scalability issue when the size of the service repository is doubled in our new benchmark. The complexity of its local search strongly depends on $n$, i.e., the dimension of each permutation. Therefore, MEGA does not even converge at all when the same amount of execution time that is required by other approaches is assigned.

Lastly, MEEDA-LOP is consistently ranked as a top performer among all the competing methods. The convergence rate of MEEDA-OP and MEEDA-TP presents a very similar pattern. However, MEEDA-OB happens to converge slower than the others, but it eventually reaches comparable results compared to  MEEDA-OP and MEEDA-TP.

\subsection{Comparison of local search operators}\label{subsection:discussion}
\begin{figure}[h!tb]
\footnotesize
\centering
\includegraphics[width=0.28\textwidth]{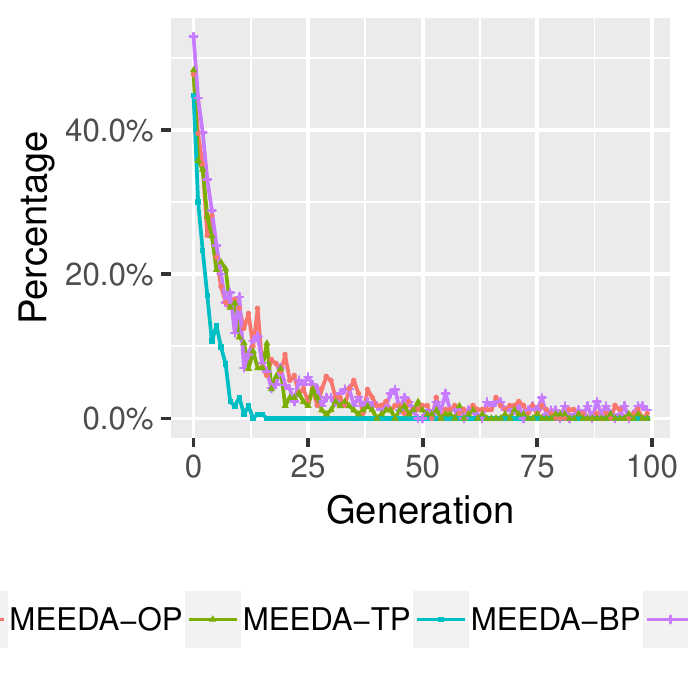}
 \caption{A comparison of the percentage of better neighbors produced by four memetic approaches along generations over 30 runs for WSC08-03}
 \label{fig:betterOP}
\end{figure}

We investigate how often the mean fitness of neighbors is better than the fitness of their original permutation in MEEDA-OP, MEEDA-TP, MEEDA-LOP, and MEEDA-BP to demonstrate which swap-based local search operator is more likely to produce better solutions. Herein we use the composition task WSC0803 as an example to demonstrate the percentage of better neighbors produced by our four memetic EDA-based approaches along generations over 30 runs for WSC08-03 in Fig.~\ref{fig:betterOP}.

The result shows that MEEDA-BP and MEEDA-TP are less like to produce better solutions while MEEDA-OP and MEEDA-LOP are very comparable to each other, although slightly higher percentages of better mean fitness can be achieved by MEEDA-LOP. 

\begin{figure}[h!tb]
\footnotesize
\centering
\includegraphics[width=0.50\textwidth]{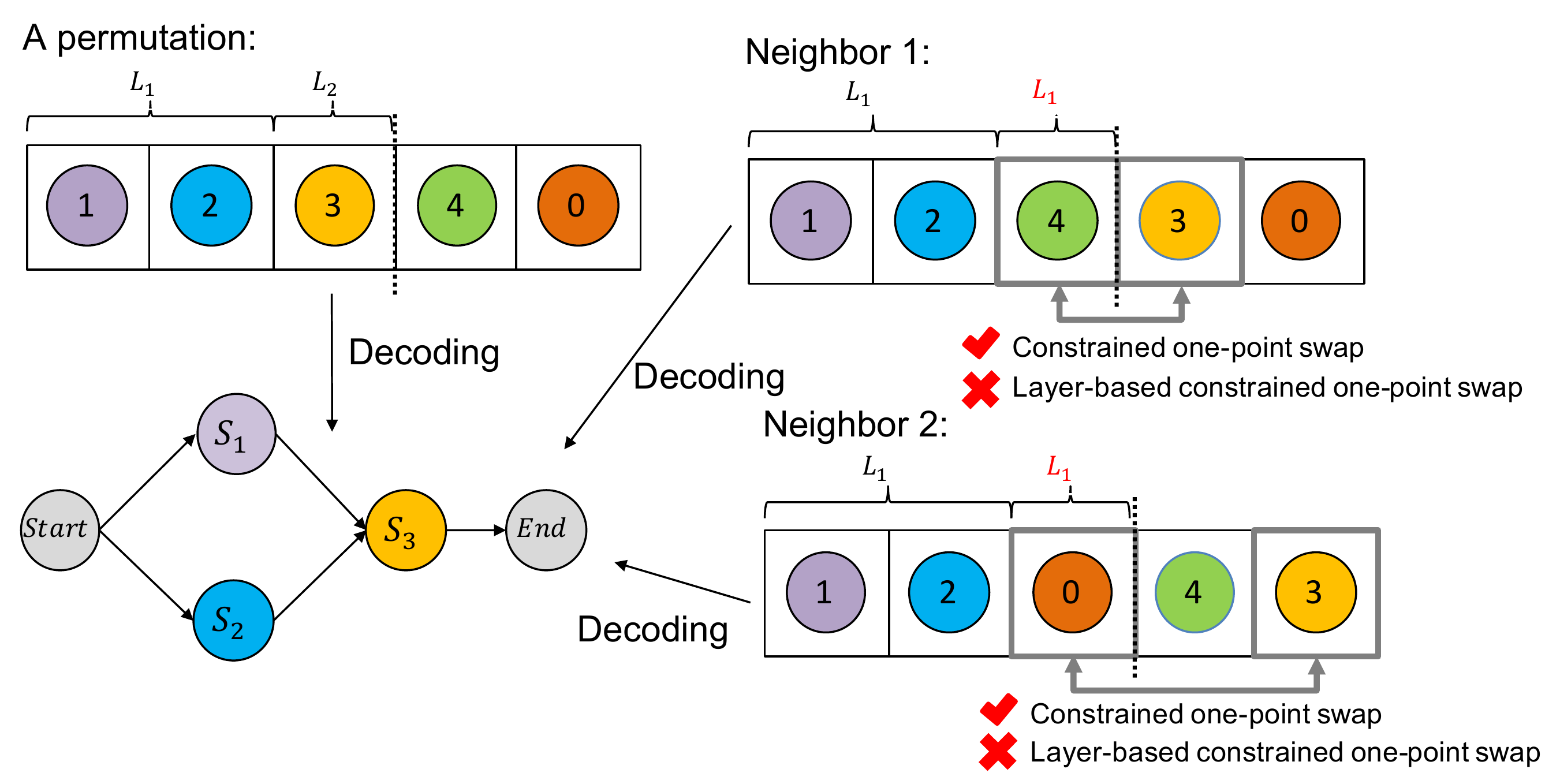}
 \caption{Examples of layer order breached by constrained one swap operation}
 \label{fig:error_swap}
\end{figure}

We further analyze differences between layer-based constrained one-point swap and constraint one-point swap operator using a permutation in Figure~\ref{fig:error_swap}.

Figure~\ref{fig:error_swap} exhibits an example of two produced neighbors from a permutation using constraint one-point swaps without considering layer information. In the example,  one identical solution can be decoded from both the given permutation and the produced two neighbors, resulting in no local exploitation. In contrast, the discussed swapping cases are not qualified for the layer-based constraint one-point swap, where any produced neighbor must strictly follow the layer order on the left-hand side of the permutation.

In the example, a given permutation is highlighted with two layers (i.e., $\mathcal{L}_1$ and $\mathcal{L}_2$) in ascending order. Particularly, $S_1, S_2 \in \mathcal{L}_1$ and $S_3 \in \mathcal{L}_2$. When the constrained one-point swap is performed,  $S_3$ in the given permutation are replaced with $S_4$ or $S_0$ in the produced neighbor 1 and neighbor 2 respectively. However, $\mathcal{L}_2$ is destroyed in the produced neighbors because of $S_4 \in \mathcal{L}_1 $ and $S_0 \in \mathcal{L}_1$.  However, if the layer-based one-point swap is applied to the given permutation, it prevents these two neighbors from being exploited. In general, all produced neighbors must keep all the ordered  layers from the given permutation.
\section{Conclusion}\label{conclusion}
In this paper, we propose effective and efficient memetic EDA-based approaches to fully automated service composition. The success of this memetic approach principally relies on the local search, where several ideas are jointly employed.  In particular,  we proposed several neighborhood structures by different local search operators, which are integrated with our permutation-based representation naturally. Besides that, a uniform distribution scheme and a stochastic strategy are also jointly utilized for selecting and applying local search. The experiments show that one of our proposed approach MEEDA-LOP achieves significantly better effectiveness and efficiency, compared to some state-of-the-art EC-based approaches and other memetic EDA-based approaches we proposed in the paper. Future work can investigate variable neighborhood with combinations of more than one local search operators in one evolutionary process, and investigate memetic EDA for handling multi-objective service composition problems.
\begin{landscape}
\begin{table}
\tiny
\centering
\caption{Mean fitness values for our approach in comparison to  NHM-EDA \cite{wang2018knowledge}, PSO \cite{wang2017comprehensive}, MEGA \cite{da2017evolutionary} and GA. (Note: the higher the fitness the better)}
\label{tbl:meanFitness}
\begin{tabular}{|c||c|c|c|c||c|c|c|c|}
  \hline
  \rule[2mm]{0pt}{0mm}
Dataset & MEEDA-OP & MEEDA-TP & MEEDA-OB & MEEDA-LOP & NHM-EDA\cite{wang2018eda} & PSO\cite{wang2017comprehensive} &MEGA \cite{da2017evolutionary} &GA\\ 
  \hline
        WSC08-1 & $\mathbf{0.613745} \pm \mathbf{0}$ & $\mathbf{0.613745} \pm \mathbf{0}$ & $\mathbf{0.613745} \pm \mathbf{0}$ & $\mathbf{0.613745} \pm \mathbf{0}$ & $0.604966 \pm 0.017232$ & $0.610182 \pm 0.003748$ & $\mathbf{0.613745 \pm 0}$ & $0.613439 \pm 0.000693$\\ 
        WSC08-2 & $\mathbf{0.756812} \pm \mathbf{0}$ & $\mathbf{0.756812} \pm \mathbf{0}$ & $\mathbf{0.756812} \pm \mathbf{0}$ & $\mathbf{0.756812} \pm \mathbf{0}$ & $\mathbf{0.756812} \pm \mathbf{0}$     & $0.756779 \pm 0.000175$ & $\mathbf{0.756812 \pm 0}$ &$\mathbf{0.756812} \pm \mathbf{0}$\\ 
        WSC08-3 & $0.477864 \pm 5.1e-05$ & $0.477866 \pm 4e-05$ & $0.477803 \pm 6.7e-05$ & $\mathbf{0.47791} \pm \mathbf{1.7e-05}$ & $0.47653 \pm 0.000212$ & $0.476086 \pm 0.000528$ & $0.477768 \pm 0.00015$ &$0.477447 \pm 0.000228$\\ 
        WSC08-4  & $\mathbf{0.557815} \pm \mathbf{0}$ & $\mathbf{0.557815} \pm \mathbf{0}$ & $\mathbf{0.557815} \pm \mathbf{0}$ & $\mathbf{0.557815} \pm \mathbf{0}$  & $\mathbf{0.557815} \pm \mathbf{0}$ & $0.557416 \pm 0.000666$ & $\mathbf{0.557815 \pm 0}$ & $0.55781 \pm 3.1e-05$\\ 
        WSC08-5  & $0.52453 \pm 0.00038$ & $0.52474 \pm 0.000388$ & $0.524482 \pm 0.000369$ & $0.524412 \pm 0.000329$ & $0.522225 \pm 0.000896$ & $0.517912 \pm 0.00544$ & $\mathbf{0.525586} \pm \mathbf{0.000639}$ & $0.523463 \pm 0.002078$\\ 
        WSC08-6  & $\mathbf{0.482698} \pm \mathbf{0.000107}$ & $\mathbf{0.482684} \pm \mathbf{0.000143}$ & $0.482622 \pm 0.000131$ & $\mathbf{0.482703} \pm \mathbf{0.000225}$ & $0.481509 \pm 0.000164$ & $0.481723 \pm 0.000505$ & $\mathbf{0.482759 \pm 0.000253}$ & $0.482142 \pm 0.000452$\\ 
        WSC08-7 & $\mathbf{0.523588} \pm \mathbf{0}$ & $\mathbf{0.523588} \pm \mathbf{0}$ & $0.523577 \pm 5.8e-05$ & $\mathbf{0.523588} \pm \mathbf{0}$ & $0.521777 \pm 0.000797$ & $0.516141 \pm 0.00519$ & $0.523559 \pm 0.00014$ &$0.521694 \pm 0.002248$\\ 
        WSC08-8 & $0.497676 \pm 1.3e-05$ & $0.497676 \pm 1.3e-05$ & $0.497646 \pm 9.4e-05$ & $\mathbf{0.497679} \pm \mathbf{0}$  & $0.491158 \pm 0.001423$ & $0.489581 \pm 0.003219$ & $0.497455 \pm 0.000357$ & $0.496347 \pm 0.000861$ \\ 
   \hline \hline
      WSC09-1  & $\mathbf{0.64902} \pm \mathbf{0.003747}$ & $\mathbf{0.650861} \pm \mathbf{0.00383}$ & $0.64852 \pm 0.003005$ & $\mathbf{0.650444} \pm \mathbf{0.00412}$ & $0.648267 \pm 0.002895$  &$0.648605 \pm 0.004328$ & $\mathbf{0.650737} \pm \mathbf{0.003863}$ & $\mathbf{0.649612} \pm \mathbf{0.003688}$ \\ 
      WSC09-2  & $\mathbf{0.521817} \pm \mathbf{0.003855}$ & $0.520562 \pm 0.004943$ & $\mathbf{0.520966} \pm \mathbf{0.008258}$ & $\mathbf{0.52309} \pm \mathbf{0.000443}$ & $0.495647 \pm 0.012517$  & $0.506701 \pm 0.011045$ & $\mathbf{0.52257} \pm \mathbf{0.003246}$ & $0.521424 \pm 0.003011$ \\ 
      WSC09-3 & $\mathbf{0.583978} \pm \mathbf{0}$ & $\mathbf{0.583978} \pm \mathbf{0}$ & $\mathbf{0.583978} \pm \mathbf{0}$ & $\mathbf{0.583978} \pm \mathbf{0}$ & $\mathbf{0.583978} \pm \mathbf{0}$  & $0.583358 \pm 0.001182$ & $\mathbf{0.583978} \pm \mathbf{0}$ & $0.583954 \pm 9.1e-05$ \\ 
      WSC09-4& $0.484428 \pm 0.000191$ & $0.484427 \pm 0.00032$ & $0.484295 \pm 0.00014$ & $\mathbf{0.485533} \pm \mathbf{0.002073}$ & $0.48222 \pm 0.000309$  & $0.481741 \pm 0.000985$ & - & $0.483277 \pm 0.000367$ \\ 
      WSC09-5 & $\mathbf{0.484832} \pm \mathbf{2.6e-05}$ & $\mathbf{0.484818} \pm \mathbf{9.5e-05}$ & $\mathbf{0.484788} \pm \mathbf{0.000179}$ & $\mathbf{0.484825} \pm \mathbf{0.000117}$ & $0.480137 \pm 0.000458$ & $0.480539 \pm 0.001308$ & $0.484603 \pm 0.000294$ & $0.483278 \pm 0.001185$ \\ 
   \hline
\end{tabular}
\tiny
\centering
\caption{Summary of statistical significance tests for fitness, where each column shows \\ win/draw/loss score of an approach against others for all instances of WSC-2008 and WSC-2009.}
\label{tbl:tableTallyFitness}
\begin{tabular}{|c@{\hspace*{5ex}}||c|c|c|c|c||c|c|c|c|}
\hline
Dataset  & Method         & MEEDA-OP & MEEDA-TP &MEEDA-OB &MEEDA-LOP & NHM-EDA\cite{wang2018knowledge}  & PSO \cite{wang2017comprehensive} &MEGA \cite{da2017evolutionary} &GA\\ 
\hline \hline
\multirow{7}{*}{\shortstack{WSC-08 \\(8 tasks)}}
                 
                          &MEEDA-OP                         
                          & -     &   0/8/0     & 0/6/2   &  \mathbf{1}/\mathbf{7}/0    &  0/2/6   &  0/1/7  & 1/4/3  &0/1/7/\\
                          &MEEDA-TP
                          &  0/8/0   & -       &  1/5/2   &  \mathbf{1}/\mathbf{6}/1     &  0/2/6   &  0/1/7  & 1/4/3 &0/1/7\\
                          &MEEDA-OB                                                
                          &  2/6/0   &  2/5/1    & -      &  \mathbf{1}/\mathbf{7}/0     &  0/2/6    &  0/1/7 & 2/5/1 &0/1/7\\
                          &MEEDA-LOP                                                
                          &  0/7/1   &  1/6/1    & 0/7/1   &-         &  0/2/6   &  0/1/7  & 1/4/3  &0/1/7\\
                          &NHM-EDA 
                         &  6/2/0   &  6/2/0    &  6/2/0   &  \mathbf{6}/\mathbf{2}/0    & -      &  0/2/6  & 6/2/0   &5/2/1\\
                          &PSO \cite{wang2017comprehensive}                                              
                          &  7/1/0   &  7/1/0    & 7/1/0   &  \mathbf{7}/\mathbf{1}/0   &  6/2/0   & -       & 8/0/0  &8/0/0\\
                          &MEGA \cite{da2017evolutionary}                                              
                         &  3/4/1   &  3/4/1    & 1/5/2   &  \mathbf{3}/\mathbf{4}/1    & 0/2/6    &  0/0/8  & -      &0/0/8\\
                         &GA                                            
                         &  7/1/0     & 7/1/0   & 7/1/0      &\mathbf{7}/\mathbf{1}/0   & 1/2/5      & 0/0/8     & 8/0/0      &-\\

\hline \hline
\multirow{7}{*}{\shortstack{WSC-09 \\(5 tasks)}}
                 
                          &MEEDA-OP                         
                          & -        &   0/5/0     & 0/4/1   &  \mathbf{1}/\mathbf{4}/0    &  0/2/3   &  0/1/4  & 0/3/1  &0/3/2/\\
                          &MEEDA-TP
                          &  0/5/0   & -           & 0/4/1   &  \mathbf{2}/\mathbf{3}/0    &  0/1/4   &  0/0/5  & 1/2/1 &0/3/2\\
                          &MEEDA-OB                                                
                          &  1/4/0   &  1/4/0    & -         &  \mathbf{2}/\mathbf{3}/0    &  0/2/3    &  0/2/3 & 1/2/1 &0/3/2\\
                          &MEEDA-LOP                                                
                          &  0/4/1   &  0/3/2    & 0/3/2     & -         &  0/1/4   &  0/1/4  & 0/3/1  &0/2/3\\
                          &NHM-EDA 
                         &  3/2/0   &  4/1/0    &  3/2/0     & \mathbf{4}/\mathbf{1}/0    & -         &  2/2/1  & 3/1/0   &2/2/1\\
                          &PSO \cite{wang2017comprehensive}                                              
                          & 4/1/0   &  5/0/0    & 3/2/0      & \mathbf{4}/\mathbf{1}/0   &  1/2/2     & -       & 3/1/0  &3/1/1\\
                          &MEGA \cite{da2017evolutionary}                                              
                         &  1/3/0   &  1/2/1    & 1/2/1      & \mathbf{1}/\mathbf{3}/0    & 0/1/3     &  0/1/3  & -      &0/2/2 \\
                         &GA                                            
                         &  2/3/0   & 2/3/0     & 2/3/0      & \mathbf{3}/\mathbf{2}/0   & 1/2/2      & 1/1/3     & 2/2/0      &-\\

\hline
\end{tabular}
\end{table}

\begin{table}
\tiny
\centering
\caption{Mean execution time (in s) for our approach in comparison to  NHM-EDA \cite{wang2018knowledge}, PSO \cite{wang2017comprehensive}, MEGA \cite{da2017evolutionary} and GA.(Note: the shorter the time the better)}
\label{tbl:meanTime}
\begin{tabular}{|c||c|c|c|c||c|c|c|c|}
\hline
\rule[2mm]{0pt}{0mm}
Dataset & MEEDA-OP & MEEDA-TP & MEEDA-OB & MEEDA-LOP & NHM-EDA \cite{wang2018knowledge}  & PSO\cite{wang2017comprehensive} &MEGA \cite{da2017evolutionary} &GA\\ 
  \hline
        WSC08-1 & $156 \pm 12$ & $211 \pm 19$ & $422 \pm 70$ & $112 \pm 8$ & $229 \pm 39$ & $\mathbf{111} \pm \mathbf{76}$ & $622 \pm 74$ &$\mathbf{102} \pm \mathbf{11}$\\ 
        WSC08-2 & $\mathbf{72} \pm \mathbf{9}$ & $98 \pm 13$ & $133 \pm 14$ & $72 \pm 6$ & $\mathbf{69} \pm \mathbf{5}$ & $68 \pm 48$ & $118 \pm 17$ & $\mathbf{21} \pm \mathbf{3}$\\ 
        WSC08-3 & $\mathbf{8470} \pm \mathbf{462}$ & $8807 \pm 621$ & $12849 \pm 740$ & $\mathbf{8329} \pm \mathbf{346}$ & $11317 \pm 7344$ & $15789 \pm 2602$ & $68382 \pm 13142$ & $12514 \pm 1575$\\ 
        WSC08-4 & $\mathbf{87} \pm \mathbf{6}$ & $112 \pm 7$ & $177 \pm 7$ & $\mathbf{84} \pm \mathbf{6}$ & $\mathbf{85} \pm \mathbf{5}$  & $211 \pm 104$ & $766 \pm 253$  & $147 \pm 36$\\ 
        WSC08-5 & $\mathbf{1705} \pm \mathbf{148}$ & $2136 \pm 142$ & $4363 \pm 154$ & $\mathbf{1742 }\pm \mathbf{122}$ & $2734 \pm 1278$ & $2549 \pm 1517$ & $47603 \pm 47104$ & $3801 \pm 1512$ \\ 
        WSC08-6 & $\mathbf{17524} \pm \mathbf{843}$ & $19954 \pm 1352$ & $43621 \pm 2062$ & $\mathbf{17303} \pm \mathbf{1569}$ & $27164 \pm 3810$ & $33119 \pm 12194$ & $947368 \pm 157828$ & $51287 \pm 11561$\\ 
        WSC08-7 & $2025 \pm 138$ & $2869 \pm 1258$ & $8096 \pm 448$ & $\mathbf{1918} \pm \mathbf{119}$  & $3993 \pm 408$  & $4456 \pm 2825$ & $81847 \pm 20610$ &$5499 \pm 1526$\\ 
        WSC08-8 & $\mathbf{4375} \pm \mathbf{371}$ & $5066 \pm 350$ & $11341 \pm 666$ & $\mathbf{4283} \pm \mathbf{368}$ & $7258 \pm 932$  & $6153 \pm 1951$ & $148133 \pm 29304$  & $10931 \pm 1667$\\ 
   \hline \hline
 	WSC09-1  & $159 \pm 23$ & $239 \pm 38$ & $314 \pm 27$ & $159 \pm 18$ & $144 \pm 93$ & $126 \pm 139$ & $506 \pm 104$ & $\mathbf{65} \pm \mathbf{12}$ \\ 
    WSC09-2  & $\mathbf{3314} \pm \mathbf{551}$ & $4311 \pm 686$ & $7573 \pm 553$ & $\mathbf{3362} \pm \mathbf{505}$ & $4642 \pm 1021$  & $\mathbf{3652} \pm \mathbf{1516}$ & $49455 \pm 17831$ & $4081 \pm 1433$ \\ 
    WSC09-3  & $\mathbf{1643} \pm \mathbf{146}$ & $2303 \pm 191$ & $4638 \pm 343$ & $\mathbf{1614} \pm \mathbf{124}$ & $\mathbf{1556} \pm \mathbf{251}$ & $\mathbf{2198} \pm \mathbf{2038}$ & $18998 \pm 3300$ & $\mathbf{1713} \pm \mathbf{394}$ \\ 
    WSC09-4  & $92342 \pm 7584$ & $103433 \pm 6847$ & $214067 \pm 12358$ & $\mathbf{86543} \pm \mathbf{6046}$ & $143098 \pm 54841$  & $\mathbf{85813} \pm \mathbf{37895}$ &- & $176152 \pm 46321$ \\ 
    WSC09-5  & $16160 \pm 1123$ & $18446 \pm 1776$ & $45039 \pm 4534$ & $\mathbf{15249} \pm \mathbf{978}$ & $26506 \pm 1716$  & $\mathbf{14807} \pm \mathbf{5605}$ & $635637 \pm 151975$ & $29991 \pm 4867$ \\ 
\hline
\end{tabular}

\tiny
\centering
\caption{Summary of statistical significance tests for execution time (in s), where each column shows \\ win/draw/loss score of an approach against others for all instances of WSC-2008 and WSC-2009.}
\label{tbl:tableTallyTime}
\begin{tabular}{|c@{\hspace*{5ex}}||c|c|c|c|c||c|c|c|c|}
\hline
Dataset  & Method & MEEDA-OP & MEEDA-TP &MEEDA-OB &MEEDA-LOP &NHM-EDA \cite{wang2018knowledge} & PSO \cite{wang2017comprehensive} &MEGA \cite{da2017evolutionary} &GA\\ 
\hline \hline
\multirow{7}{*}{\shortstack{WSC-08 \\(8 tasks)}}
                          &MEEDA-OP                         
                          & -     &  0/0/8     &  0/0/8   &  \mathbf{6}/\mathbf{2}/0    &  0/2/6   &  1/1/6  & 0/0/8   & 2/0/6 \\
                          &MEEDA-TP
                         &  8/0/0   & -       &  0/0/8   &  \mathbf{8}/0/0    &  2/1/5   &  2/1/5   & 0/0/8   & 2/0/6 \\
                          &MEEDA-OB                                                
                         &  8/0/0   &  8/0/0    & -      &  \mathbf{8}/0/0     &  7/1/0   &  5/1/2  & 0/0/8   &  4/3/1  \\
                          &MEEDA-LOP                                                
                           &  0/2/6   &  0/0/8    &  0/0/8   &-       &  0/0/8   &  0/2/6  & 0/0/8    & 2/0/6 \\
                          &NHM-EDA 
                         &  6/2/0   &  5/1/2    &  0/1/7   &  \mathbf{8}/0/0     & -      &  2/4/2  & 0/0/8   &  2/1/5 \\
                          &PSO \cite{wang2017comprehensive}                                              
                         &  6/1/1   &  5/1/2    &  2/1/5   & \mathbf{6}/\mathbf{2}/0    &  2/4/2   &-        & 0/0/8   &  2/4/2 \\
                          &MEGA \cite{da2017evolutionary}                                              
                          &  8/0/0   &  8/0/0    &  8/0/0   & \mathbf{8}/0/0   & 8/0/0    &8/0/0    &-        &  8/0/0 \\
                         &GA                                             
                         &  6/0/2     & 6/0/2       & 1/3/4      &\mathbf{6}/0/2  & 5/1/2      &  2/4/2     & 0/0/8      &- \\
\hline \hline
\multirow{7}{*}{\shortstack{WSC-09 \\(5 tasks)}}
                          &MEEDA-OP                         
                          & -       &  0/0/5     &  0/0/5   &  \mathbf{2}/\mathbf{3}/0     &  0/2/3   &  0/1/4   & 0/0/4   & 3/1/1 \\
                          &MEEDA-TP
                         &  5/0/0   & -          &  0/0/5   &  \mathbf{5}/0/0     &  2/1/2   &  4/1/0   & 0/0/4   & 4/1/0 \\
                          &MEEDA-OB                                                
                         &  5/0/0   &  5/0/0    & -         &  \mathbf{5}/0/0     &  5/0/0   &  5/0/0  & 0/0/4   &  5/0/0  \\
                          &MEEDA-LOP                                                
                         &  0/3/2   &  0/0/5    &  0/0/5    &-           &  0/2/3   &  0/5/0  & 0/0/4   & 1/1/3 \\
                          &NHM-EDA  
                         &  3/2/0   &  2/1/2    &  0/0/5    &  \mathbf{3}/\mathbf{2}/0     & -        &  3/2/0  & 0/0/4   &  1/2/2 \\
                          &PSO \cite{wang2017comprehensive}                                              
                         &  4/1/1   &  0/1/4    &  0/0/5    &  0/\mathbf{5}/0     &  0/2/3   &-        & 0/0/4   &  1/2/2 \\
                          &MEGA \cite{da2017evolutionary}                                              
                         &  4/0/0   &  4/0/0    &  4/0/0    & \mathbf{4}/0/0      & 4/0/0    & 4/0/0    &-        &  4/0/0 \\
                         &GA                                             
                         &  1/1/3   &  0/1/4    &  0/0/5    & \mathbf{3}/\mathbf{1}/1      & 2/2/1    & 2/2/1   & 0/0/4     &- \\

\hline
\end{tabular}
\end{table}
\end{landscape}

\bibliographystyle{IEEEtranBib}
\bibliography{IEEEexample} 
\end{document}